\documentclass[journal]{IEEEtran}
\usepackage{times}
\usepackage{epsfig}
\usepackage{graphicx}
\usepackage{graphics}
\usepackage{epstopdf}
\usepackage{subfigure}
\usepackage{amsmath}
\usepackage{amssymb}
\usepackage{cite}
\usepackage{booktabs}
\usepackage{url}
\usepackage{gensymb}
\usepackage[colorlinks]{hyperref}
\begin{document}

\ifCLASSINFOpdf
\else

\fi

\hyphenation{op-tical net-works semi-conduc-tor}

\title{Multi-Person Pose Estimation with Enhanced Feature Aggregation and Fusion\\}

\author{
        Xixia~Xu,
        Qi Zou$^*$\thanks{$^*$ Corresponding author},
        Xue~Lin
        \thanks{Xixia Xu, Qi Zou and Xue Lin are with School of Computer and Information Technology, Beijing Jiaotong University, Beijing 100044, China (email : 18120432@bjtu.edu.cn; qzou@bjtu.edu.cn; 18112028@bjtu.edu.cn).}
        }

        % <-this % stops a space

\markboth{Multi-Person Pose Estimation with Enhanced Feature Aggregation and Fusion}%
{Shell \MakeLowercase{\textit{et al.}}: Bare Demo of IEEEtran.cls for IEEE Journals}
%\markright{IEEE TRANSACTIONS ON IMAGE PROCESSING}
\maketitle

\begin{abstract}
  We propose a novel Enhanced Feature Aggregation and Fusion network (EFAFNet) for multi-person 2D human pose estimation. Due to enhanced feature representation, our method can well handle crowded, cluttered and occluded scenes. More specifically, a Feature Aggregation and Selection Module (FASM), which constructs hierarchical multi-scale feature aggregation and makes the aggregated features discriminative within one single block, is proposed to get more accurate fine-grained representation, leading to more precise joint locations. Then, we perform a simple Feature Fusion (FF) strategy which effectively fuses high-resolution spatial features and low-resolution semantic features to obtain more reliable context information for well-estimated joints. Finally, we build a Dense Upsampling Convolution (DUC) module to generate more precise prediction, which can recover missing joint details that are usually unavailable in common upsampling process and to reduce the false joint information. As a result, the predicted keypoint heatmaps are more accurate. Comprehensive experiments demonstrate that the proposed approach outperforms the state-of-the-art methods and achieves the superior performance over three benchmark datasets: the recent big dataset CrowdPose, the COCO keypoint detection dataset and the MPII Human Pose dataset. Our code will be released once the paper is published.
\end{abstract}


\IEEEpeerreviewmaketitle

\section{Introduction}

\IEEEPARstart{M}{ulti-person} Pose Estimation devotes to locate body parts for multiple persons in an image, such as keypoints on the arms, torsos, and the face\cite{li2019multi-person,ning2018knowledge-guided}. The related tasks contain pose estimation in videos\cite{zhou2016spatio-temporal} and 3D pose estimation. It regards as a fundamental task to deal with other related high level tasks, such as human-computer interaction\cite{marcosramiro2015let}, human action recognition\cite{ZhaoxuanAttention,cai2016effective}, emotion analysis\cite{Buitelaar2018MixedEmotions} and motion capture\cite{kadu2014automatic}, content retrieval\cite{ren2012visual}, human performance analysis\cite{torres2018a}.

Recently, due to the rapid development of convolution neural networks (CNN)\cite{he2016deep}, most existing methods\cite{chen2018cascaded,fang2017rmpe:,li2018crowdpose:,2019Pose,ke2018multi-scale,xiao2018simple} have achieved remarkable advances in multi-person pose estimation.
\begin{figure}[htb]
\centering
\includegraphics[width=2.7in,height = 3.0in]{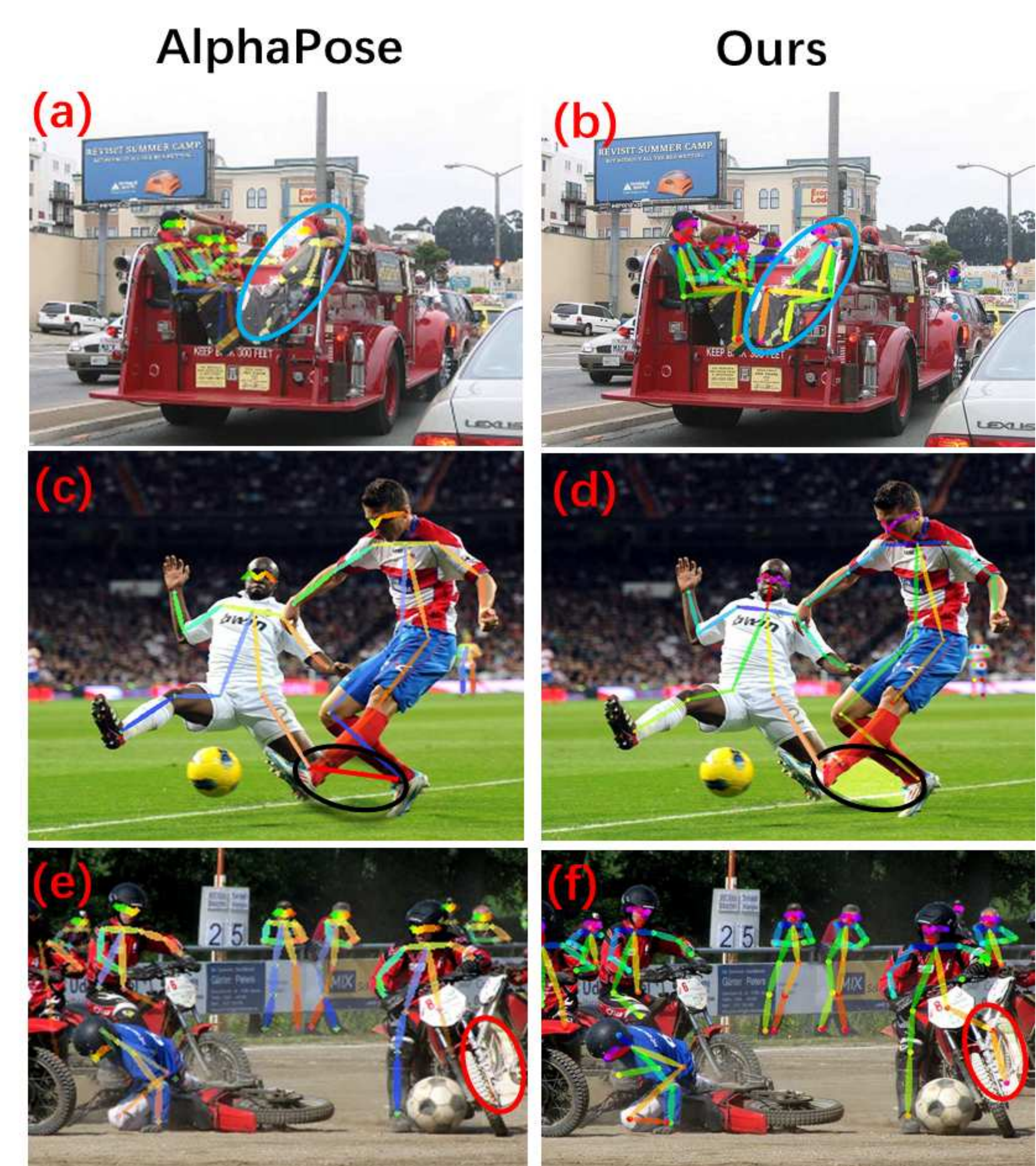}
\caption{Qualitative comparison of AlphaPose\cite{fang2017rmpe:} and our method in complex scenes. The major difference between the two methods is marked with a circle.}
\label{fig:1duibi1}
\end{figure}
They can be roughly classified into two frameworks, i.e., the bottom-up framework\cite{cao2017realtime,insafutdinov2016deepercut:,newell2017associative} and the top-down framework\cite{chen2018cascaded,fang2017rmpe:,li2018crowdpose:,2019Pose}. The former first detects all body keypoints together and then assigns the detected joints to different human instances. In contrast, the latter one detects all human bounding boxes in one image at first and then estimates the pose according the detected human proposal. Besides, other methods do exist, e.g., \cite{li2019multi-person} can solve the pose estimation by combining the top-down and bottom-up method into a single framework. With the prior information of human detection, top-down methods can get more reliable and accurate estimations. Our model belongs to the top-down framework.

However, despite the current methods have achieved promising results on standard benchmarks such as MPII Human Pose \cite{andriluka20142d} and COCO keypoint detection dataset\cite{lin2014microsoft}, some challenges have not been well addressed. One of them is the missing detection of joints due to cluttered background like Fig.\ref{fig:1duibi1} (a), humans in multiple scales, or due to occlusions like Fig.\ref{fig:1duibi1} (e). Another is false assembling of joints into a pose due to overlapping in crowded scenes like Fig.\ref{fig:1duibi1} (c).

The main reasons lie in several points: $\textbf{1)}$ multi-scale features are not fully exploited, so that some methods are less robust to pose variations and less adaptive to multi-scale persons. Most existing methods only represent the multi-scale features in a coarse-grained layer-level, but neglect the fine-grained multi-scale information within each layer.
$\textbf{2)}$ Most CNN-based methods treat the contributions of different joint locations and multiple channels equally, which results in lacking of discriminative spatial information and semantic information across channels for pose estimations.
$\textbf{3)}$ Context information is not well represented in many existing methods, which results in failing to infer the invisible or indistinguishable joints, such as the left knee and ankle of the man on the motorbike in (Fig.\ref{fig:1duibi1} e), or wrong assembling of joints into a pose. $\textbf{4)}$ Missing detailed information in the downsampling process is not effectively recovered in most methods, which makes models tend to get the false or missing joints in the final output. Besides, computational efficiency is also a problem. Some existing methods cost a large number of computing resources and design complex networks to address these problems.

To handle the above problems effectively and efficiently, we propose an Enhanced Feature Aggregation and Fusion network (EFAFNet), as shown in Fig.\ref{fig:2network}. Based on the ResNet\cite{he2016deep} backbone, the whole framework consists of two part: encoder and decoder network. In the encoder, a $\emph{Feature Aggregation and Selection Module (FASM)}$ is added in the bottleneck to enhance the multi-scale feature extractions and adaptively select the most discriminative human part regions. Before start the process of decoder, we combine the low-level spatial information with the high-level semantic concepts in the $\emph{Feature Fusion Module (FFM)}$. Based on the fused features, we attach a $\emph{Dense Upsampling Convolution (DUC)}$ module in the decoder to explicitly recover the detailed keypoint information.

Our network has several benefits in comparison to existing widely-used methods for pose estimation. On the one hand, $\emph{Feature Aggregation Module (FAM)}$ extracts features in group-wise manner rather than in layer-wise manner. Therefore it can aggregate stronger features using a smaller number of parameters. On the other hand, $\emph{Feature Selection Module (FSM)}$ can obtain the enhanced discriminative feature representations, and accordingly the predicted heatmap is spatially more precise. Additionally, most existing fusion schemes aggregate low-level and high-level representations using the dense connection and cost a large number of the computer resources. Instead, we combine the low-level structural features and the high-level semantic information in an efficient way to capture rich context information adequately. Consequently, we can better infer the invisible or occluded joint with this context information and reduce the wrong assembling of joints. Inspired by work in semantic segmentation\cite{wang2018understanding}, we build a $\emph{DUC}$ module on head of network rather than the common bilinear upsampling, which can decode more detailed information which are missing in the downsampling process. It can naturally fit the overall framework by training in an end-to-end manner, and it increases the AP on the COCO dataset\cite{lin2014microsoft} significantly, especially on relatively small persons.

In summary, our main contributions are four-fold as follows:

$\bullet$ We propose a Feature Aggregation and Selection Module (FASM). The Feature Aggregation Module (FAM) captures the multi-scale fine-grained information for body joints in an efficient way, while the Feature Selection Module (FSM) adaptively exploits more discriminative information of body parts.

$\bullet$ We adopt a simple Feature Fusion (FF) strategy, which lets spatial high-resolution information and semantic concepts interact and benefit each other, to effectively integrate and exploit global contextual information that benefits to infer occluded body parts.

$\bullet$ We build a Dense Upsampling Convolution (DUC) module to reduce the false joint information and to recover missing joint details that are usually unavailable in common upsampling process.

$\bullet$ Our method achieves superior performance and high efficiency on the COCO keypoint dataset\cite{lin2014microsoft} and MPII human pose benchmark\cite{andriluka20142d} and achieves the comparable results on the big CrowdPose dataset\cite{li2018crowdpose:} without using extra data.

\section{Related Work}
Recently, human pose estimation has been became more and more popular. Recent works\cite{jain2014learning,newell2016stacked,tompson2014joint,wei2016convolutional,bulat2016human} mostly rely on the development of CNN\cite{he2016deep}, which progressively improves the performance of pose estimation. The topic is categorized into two parts. The former is the single-person pose estimation that predicts a single person keypoints based on the cropped image patches given the detected bounding box. The latter is the multi-person pose estimation that requires further localization of the human keypoints of all persons in one image.

Our proposed approach is concerned about previous works referring to multi-person pose estimation, as described as follows:

\begin{figure*}[htb]
\centering
\includegraphics[width=6.42in,height=2.89in]{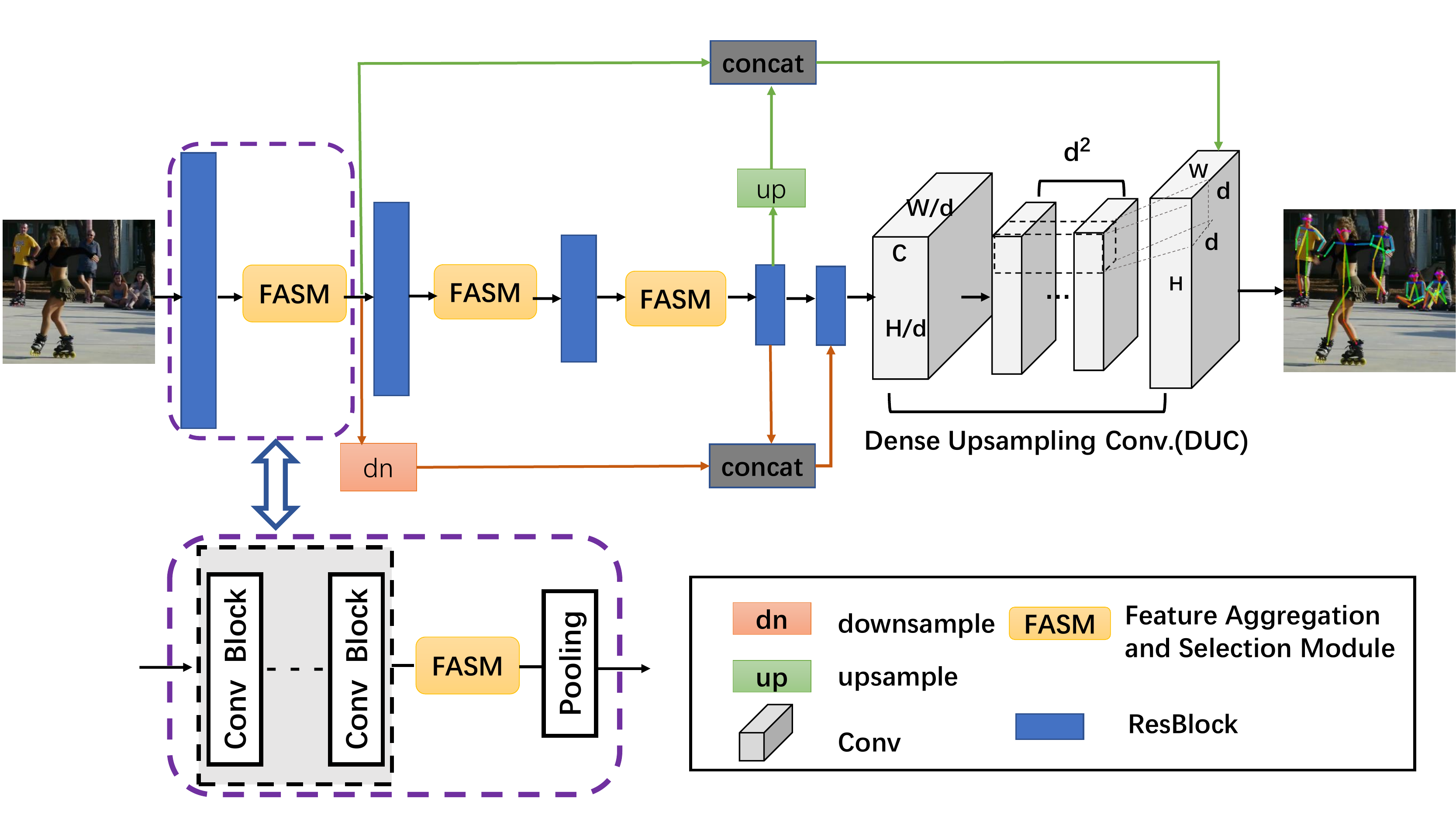}
\caption{Overview of the proposed EFAFNet for multi-person pose estimation. The proposed FASM is placed at every bottleneck of the encoder and regards as input features extracted from the convBlock of ResBlock\cite{he2016deep}. It is depicted in a purple dashed line box. The feature map from first resblock (green lines) concatenates with feature map from the last resblock (pink lines). The dense upsampling convolution module is placed at the head of the network to decode the output feature maps to the stable resolution which recovers rich and detailed contextual information.}
\label{fig:2network}
\end{figure*}

%------------------------------------------------------------------------
%-------------------------------------------------------------------------

\subsection{Multi-Person Pose Estimation}
Recently, multi-person pose estimation is gaining more and more popular due to the real-life demand. There are two categories in the multi-person pose estimation. The first one is bottom-up approach, which obtains all the body keypoints in the input image and assembles the detected keypoints into different person using their methods. The second one is top-down approach, which firstly uses human detector to get bounding boxes and feed the cropped image to the pose estimation network.

\textbf{Bottom-up.} Bottom-up approaches\cite{cao2017realtime,insafutdinov2016deepercut:,newell2017associative,pishchulin2016deepcut:} firstly detect the human body poses and then group them with the clustering algorithm to different human instances. Compared with the top-down approaches, they are faster in testing and lighter in building model. However, the bottom-up method loses the chance to amplify the details of each person, this results in its accuracy lower than the top-down approaches. DeepCut\cite{pishchulin2016deepcut:} regards the keypoints localization as an Integer Linear Program (ILP) problem and assigns keypoint into different person groups. After that we combine the person clusters with corresponding body parts to obtain the final results. DeeperCut\cite{insafutdinov2016deepercut:} improves DeepCut\cite{pishchulin2016deepcut:} using ResNet\cite{he2016deep} to extract more strong body parts representations and adopts image-conditioned pairwise terms to achieve better performance.
Associative Embedding\cite{newell2017associative}produces confidence maps and pixel-wise embedding in the meantime and group the keypoint candidates into different instances to generate the final result. MultiPoseNet\cite{kocabas2018multiposenet:} jointly localizes the human detection and multi-person pose estimation, which designs a pose residual network to receive keypoints and person detections and obtain accurate estimation result by assigning keypoints to different person using the PRN module.

\textbf {Top-Down.}
Top-down approaches\cite{fang2017rmpe:,xiao2018simple,ning2018knowledge-guided,papandreou2017towards,chen2018cascaded} regard the process of predicting keypoints as a two-step operation. Actually, top-down methods firstly detect all the persons in an image and crop the person region, then feed the cropped image into the single person pose estimation model. G-RMI\cite{papandreou2017towards} predicts heatmaps and offsets of the points on the heatmaps to the labeled heatmaps, and then get the final location with the heatmaps and offsets. Moreover, the Cascaded Pyramid Network\cite{chen2018cascaded} divides keypoints into the hard and easy level. It uses the feature pyramid architecture\cite{lin2017feature} as backbone and divides the whole framework into the GlobalNet and RefineNet, to locate the easy and hard keypoints respectively. Our work follows the two-step framework\cite{xiao2018simple} which based on a simple encoder-decoder architecture and adopts a few transposed convolution layers for generating high-resolution representations. The top-down framework can achieve superior performance as the object detection and single person pose estimation develop.
%-------------------------------------------------------------------------
\subsection{Single Person Pose Estimation}

It is much simpler than multi-person pose estimation and only needs to obtain the body parts of a single person that is already cropped in an image. Conventional methods for single-person pose estimation are mostly based on pictorial structure
models\cite{pishchulin2013poselet,sapp2013modec:,sun2011articulated,tian2012exploring}.
Since the DeepPose\cite{toshev2014deeppose:} research, human pose estimation study has converted from traditional methods to deep learning stage because of their superior performance. Recent methods\cite{tang2018deeply,sun2017human,ke2018multi-scale} have achieved promising performance on recent popular datasets\cite{andriluka20142d,lin2014microsoft}. For single person pose estimation, their performance are more rely on the person detector.

\subsection{Multi-scale Feature Representations and Fusions.}
From the traditional feature design\cite{lowe2004distinctive} to deep learning\cite{hou2017deeply,szegedy2015going}, the multi-scale features have been widely used in many research. A large receptive field is necessary for obtaining multi-scale feature representations to locate joints at different scales precisely. In the previous works for the multi-person pose estimation, the large receptive field is achieved by the architecture in the Convolutional Pose Machines\cite{wei2016convolutional} to implicitly capture the spatial relations among different parts, resulting in the progressively refined estimations. However, low-level information is not considered well. Stacked Hourglass Networks\cite{newell2016stacked} process the feature maps of all different scales to capture various spatial relationships of different resolutions, and adopt the skip connections to save information at each resolution. But this model works at a cost of a large amount of the computing resource and may not dig out the information exhaustively. By the above observation, we propose the $\emph{FAM}$ to aggregate the features at a granular and lightweight level in the bottleneck and amplify the receptive fields range for each network layer.

The multi-level feature fusion is crucial for the keypoint localization as well. For example, Stacked Hourglass Networks\cite{newell2016stacked} fuse the different resolution feature maps in a dense way and achieve a good performance. The Cascaded Pyramid Network\cite{chen2018cascaded} preserve the high-level and low-level information from the feature maps across different scales to obtain more information. However, the above methods fuse the different level features with connections in complex forms and concepts. And it's still hard to make a better result. Thus, we propose a simple $\emph{FFM}$ to combine the high-level semantic information with low-level features to enhance the global context information.

\subsection{Decoding of Feature Representation.}
The downsampling operations in CNN architectures lead to the essential fine details (low-level information) missing, which cannot be well recovered by the common up-sampling operations. There are many methods propose to decode detailed information which are generally lost in the downsampling process. Bilinear upsampling operation is commonly used\cite{chen2018deeplab:,zeiler2014visualizing}, due to its fast speed and efficient memory. Another common used operation is deconvolution, in which the unpooling operation use the stored pooling shifted from the pooling to recover the essential information for visualization\cite{zeiler2014visualizing}. As the FCN\cite{long2015fully} illustrates, in the decoding stage place the deconvolutional layer to generate result with the help of the cascaded intermediate feature maps. DeconNet\cite{noh2015learning} adopt deconvolutional layers in unpooling operation by using the preserved pooled location. By the above observation, we find that there are few existing methods pay attention to the inevitable information missing during the encoding stage for the keypoint localization. Except that, simple baseline\cite{xiao2018simple} method adopts a few transposed convolution layers for generating high-resolution representations, which are vital to recover more fine detailed information than the bilinear interpolation. However, we think it still can result in some detailed information missing, so we propose to use the $\emph{DUC}$\cite{wang2018understanding} module to replace the deconvolutional layers to make up for the missing details and obtain the final prediction. Different from the recent HRNet\cite{sun2019deep}, it connects high-to-low resolution sub-networks in parallel and maintains the same high-resolution horizontally so avoids the information loss resulting from the upsampling. Our DUC module is mainly designed for the most existing pose estimation methods that adopt the up-sampling operations.

\section{Method}
In this section, we propose a novel EFAFNet to make the localization more precisely. It can not only make use of the context information fully but extract the rich discriminative multi-scale features. In addition, the model can recover more detailed information that is mostly missing in the downsampling process. An overview of our proposed framework is illustrated in Fig.\ref{fig:2network}. Firstly, we briefly review the Simple Baseline Network(SBN)\cite{xiao2018simple} framework. After that, we introduce the FASM, FFM and DUC module in detail.

\subsection{Review Simple Baseline Network}
Simple Baseline Network (SBN)\cite{xiao2018simple} adopts three deconvolution layers after the last convolution stage of the ResNet\cite{he2016deep}, in which each deconvolution layer has $256$ filters with $4$ $\times$ $4$ kernel size and stride is $2$. Finally, a $1$ $\times$ $1$ convolution layer is added to the deconvolution module to predict the keypoint heatmaps.

\subsection{FASM: Feature Aggregation and Selection Module}
\begin{figure}[h]
\centering
\includegraphics[width=3.3in,height = 3.6in]{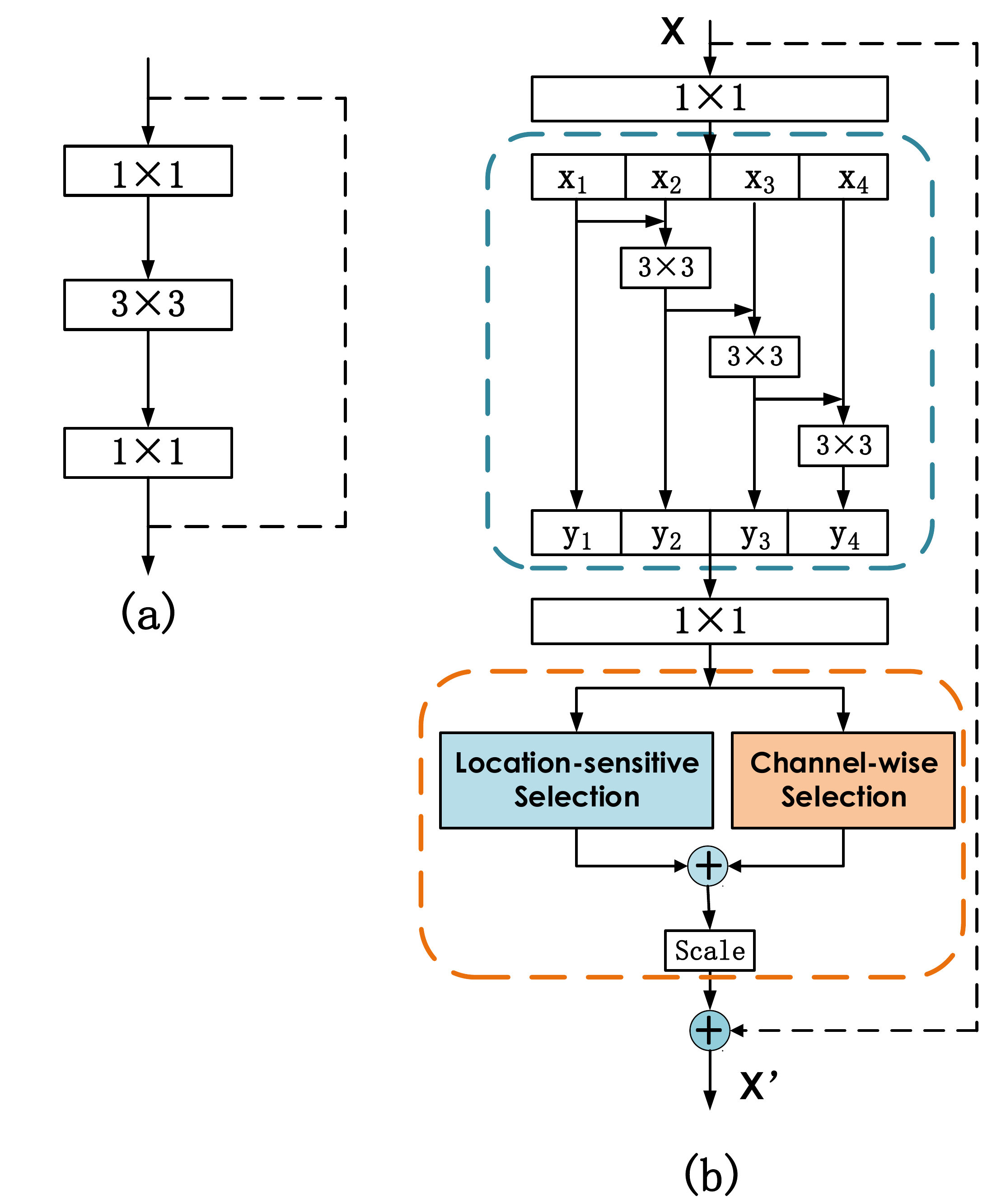}
\caption{The schema of the Bottleneck (a) and Feature Aggregation and Selection Module(FASM) (b) which is composed of the Feature Aggregation Module (the blue rectangle enclosed) and Feature Selection Module (the red rectangle enclosed).}
\label{fig:3 FASM}
\end{figure}

In order to extract informational and discriminative feature of human parts, we design a Feature Aggregation and Selection Module in each bottleneck block as shown in Fig.\ref{fig:3 FASM}. FASM consists of FAM and FSM. The FAM extracts more multi-scale features to enhance the feature representation ability of the human body part information. FSM is followed by the FAM operation, which is mainly used for adaptively highlighting the discriminative joint features among different feature maps and different locations. The detailed explanations are discussed in followings.

\subsubsection{Feature Aggregation Module}
Since human pose estimation is a typical structure prediction problem, it is essential to explore both local characteristics (such as some area around the torsos) and global context information, which is benefit to model the spatial relationship between joints. Accordingly, a larger receptive field of network is preferable. To realize the goals, FAM is proposed to extract robust features for multi-scale stimuli for the network.
Superior to most existing methods that only consider coarse multi-scale information across different layers of CNN frameworks, we also capture fine-grained multi-scale information within each layer in a lightweight way.

In each resblock, bottleneck structure\cite{he2016deep} is applied, which is a populer building block in many modern backbone CNNs architectures, as shown in the Fig.\ref{fig:3 FASM} (a). Our proposed multi-scale FAM implements based on it. As shown in the blue rectangle enclosed by dashed line at the top of the Fig.\ref{fig:3 FASM} (b). After the $1$ $\times$ $1$ convolution, we evenly split the feature maps into $4$ feature map subsets, denoted
by $x$$_{i}$, where $i$$\in$ {$1$, $2$, $3$, $4$}. Compared to the input feature map, each sub-map $x$$_{i}$ has the same spatial size but the $1$/$4$ number of channels. Besides $x$$_{1}$, each $x$$_{i}$ is plunged into a corresponding $3$ $\times$ $3$ convolution \textbf{G$_{i}$}(), whose output is denoted as $y$$_{i}$ $=$ \textbf{G$_{i}$}($x$$_{i}$). To characterize the fine-grained multi-scale information within each layer, we connect the output of each sub-map in a residual-like manner. The feature subset $x$$_{i}$ is added with the output of \textbf{G$_{i-1}$}(), and then fed into \textbf{G$_{i}$}. If we want to reduce parameters, the subset numbers will increase. Thus, we don't apply the $3$ $\times$ $3$ convolution for $x$$_{1}$. Thus, $y$$_{i}$ can be represented as:

\begin{equation}
\label{eq6}
y_{i}=\left\{
\begin{aligned}
x_{i} & , & i=1, \\
G_{i}(x_{i}+y_{i-1}) & , & 1< i\leq s.
\end{aligned}
\right.
\end{equation}

Intuitively, benefit from the connections between sub-maps, each $3$ $\times$ $3$ convolutional operator \textbf{G$_{i}$}() could potentially receive feature information from all feature splits {$x$$_{j}$, $j$$\leq$ $i$}. Furthermore, the obtained feature $y$$_{i}$ will correspond to a larger receptive field than the feature extracted by standard bottleneck structure without FAM. In summary, FAM is effective to obtain multi-scale features of detected persons, which takes full advantages of local information of joints and global relationship of different parts among human. In order to fuse information across different scales fully, we concatenate all branches and feed them into a $1$ $\times$ $1$ convolution. The split and concatenation operation can use convolutions to extract more robust features effectively. This module can reduce the parameters of the whole model on the basis of accurate localization.

\subsubsection{Feature Selection Module}
For the multi-person pose estimation, the extracted human features about the locations and semantics are possibly redundant. It's not effective for treating the contributions of different joint spatial locations and channel information across feature maps equally. The different keypoint locations corrsponding to different spatial information for the human part. The different channels among different feature maps related to the multi-level semantic information. In order to highlight the discriminative information both in the spatial location and channel context, we propose FSM, which aims to learn location-sensitive weights $\beta$ and channel-wise weights $\alpha$ for each feature map respectively.
The detailed design of FSM is shown as Fig.\ref{fig:3 FASM}, the bottom half of the figure enclosed by the red rectangle dashed line.

\textbf{location-sensitive selection.}
Applying the whole feature maps may lead to sub-optimal results due to the irrelevant regions. Different from paying attention to the whole image region equally, $\emph{Location-Sensitive Selection (LSS)}$ mechanism attempts to adaptively highlight the part-related regions in the feature maps.

Assuming the input of the LSS block is $\textbf{X}\in\mathbb{R}^{C \times H \times W}$ (the output of FAM), and the output of the location-sensitive selection block is the location-sensitive weight $\beta\in\mathbb{R}^{H\times W}$, the final discriminative enhanced feature maps mapping to the human part location are $\textbf{X}^\prime\in\mathbb{R}^{C \times H \times W}$. The location-sensitive weight $\beta$ is generated by one convolution operation $\textbf{W$_{1}$}\in\mathbb{R}^{1 \times 1 \times C}$ and another convolution operation $\textbf{W$_{2}$}\in\mathbb{R}^{1 \times 1 \times 1}$ , a relu operation followed by a sigmoid function in the input $\textbf{X}$, i.e.,
\begin{equation}
\beta = \emph{Sigmoid}(\sigma(W_{2}(\sigma(W_{1}(\emph{X})))))
\end{equation}
where $\textbf{W}$ denotes the convolution weights, the $\textbf{$\sigma$}$ represents the relu activation function and $\emph{Sigmoid}$ means the sigmoid activation function.

Finally, the learned location-sensitive weights $\beta$ is rescaled on the input $\textbf{X}$ and then add back to the input to achieve the final enhanced output $\textbf{X}^\prime$.
\begin{equation}
x'_{i,j} = x_{i,j}+\beta_{i,j}\otimes x_{i,j},
\end{equation}
where $\otimes$ means the element-wise multiplication between the
\emph{i, j}-th element of $\beta$ and $\textbf{X}$ in the joint location context.

\textbf{channel-wise selection.}
Since convolutional filters operate as the pattern detector. After the convolutional operation, each channel of a feature map equals to the feature responses of the corresponding filter.
For the keypoint localization, different channels of features in CNNs generate different response to different parts among the whole body. This selection mechanism produce a channel-wise vector by exploiting the inter-channel relationship of features.
It can be viewed as a process of adaptively selecting the significant information for the human among different feature maps. After that, we can guide the network to eliminate the false results with the help of the selected discriminative information.

\begin{figure}[h]
\centering
\includegraphics[width=3.2in,height = 2.0in]{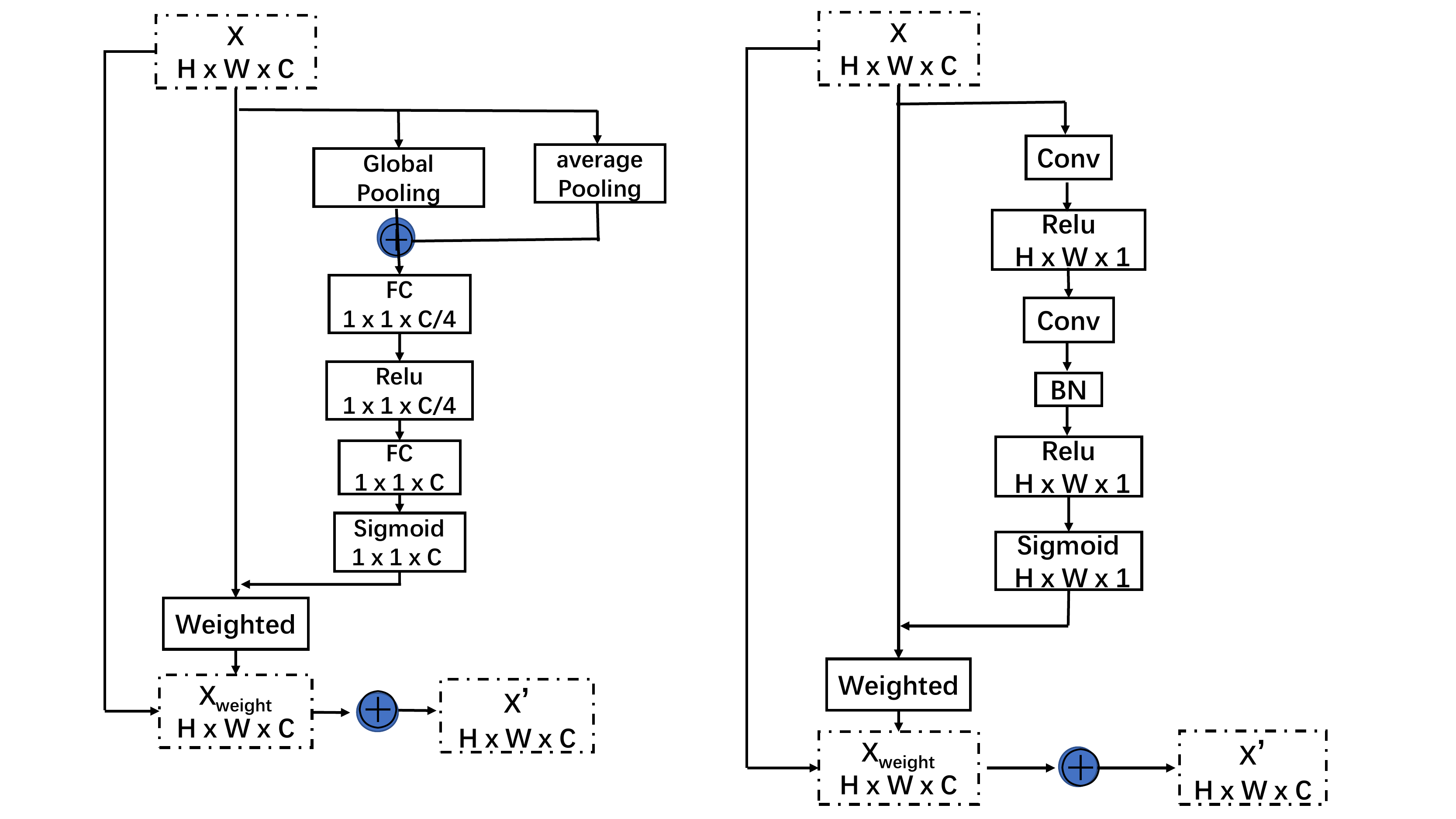}

\caption{Channel-wise Selection (left) and Location-Sensitive Selection(right). Where X and X' mean weighted feature and final enhanced feature respectively, X$_{weight}$ means weighting feature in this paper.}

\label{fig:4 channelandspatial}
\end{figure}

The $\emph{Channel-wise Selection (CS)}$ module will assign larger weight to channels which show high response to the human body part. As shown in Fig.\ref{fig:4 channelandspatial}.

To aggregate the feature map for each channel, we take the $\textbf{X}$ (the output of FAM) as the input of the channel-wise selection and the output of the channel-wise selection vector $\alpha\in\mathbb{R}^{C}$, this vector softly encodes the global and local information for different feature maps. Thus we acquire the informative features $\textbf{X}^\prime\in\mathbb{R}^{C \times H \times W}$  among the channel-wise context.

The CS weight matrix is computed in two steps analogous to the SE-Net\cite{hu2018squeeze-and-excitation} but is different from that. We observe that the max-pooling operation can extract the distinctive information to infer detailed channel-wise highlights. Thus, we apply both average-pooling and max-pooling features simultaneously. Firstly, global average pooling is performed on the input $\textbf{X}$ to generate the global statistics \textbf{X$_{ave}$} and then the max pooling operation do the same operation we get the local statistics \textbf{X$_{max}$}. We combine them in the form of element-wise add and concatenate respectively, and by comparing the experimental results, we choose to use element-wise add finally.

It is just as described as follows:
\begin{equation}
X = X_{avg} \textbf{$\oplus$} X_{max},
\end{equation}
where the $\textbf{$\oplus$}$ is the element-wise add operation.

The channel-wise vector $\emph{z}\in\mathbb{R}^{C}$, where the $\emph{c}$-th element of $\emph{z}$ is obtained by
\begin{equation}
\emph{z}_\emph{c} =\frac{1}{H\times W}\sum_{i=1}^{H}\sum_{j=1}^{W}x_{c}(i,j),
\end{equation}
where $b_{c}\in\mathbb{R}^{H \times W}$ is the $\emph{c}$-th element of the input $\textbf{X}$.

In the next step, a selecting mechanism with a sigmoid activation is acted on the channel-wise statistics $\emph{z}$, i.e.,
\begin{equation}
\alpha = \emph{Sigmoid}(W_{2}(\sigma(W_{1}(\emph{z})))),
\end{equation}
where $\textbf{W$_{1}$}\in\mathbb{R}^{C\times C}$ and $\textbf{W$_{2}$}\in\mathbb{R}^{C\times C}$ represents two fully connected layers, $\sigma$ means the ReLU activation function.

Finally, the learned CS weights $\alpha$ is
rescaled on the input $\textbf{X}$ and then add back to the input to achieve the final output
$\textbf{X}^\prime$, i.e.,
\begin{equation}
x'_{c} = x_{c}+\alpha_{c} \otimes x_{c},
\end{equation}
where $\otimes$ means the element-wise multiplication between the
$\emph{c}$-th element of $\alpha$ and $\textbf{X}$ in the channel-wise global context.

Some examples can reflect the FSM refinement process are shown in Fig.\ref{fig:8FSM map}. We can see false alarms for different body parts are reduced and heatmaps are corrected by selecting the significant human body part regions. The FSM can correct the initial estimation of the occluded or similar joints relying on the discriminative location and channel-wise context information.

\subsubsection{Arrangement of the CS and LSS module}

As shown in Fig.\ref{fig:3 FASM}, assuming the input of residual bottleneck is $\textbf{X}\in\mathbb{R}^{C \times H \times W}$, the FSM is performed on the non-identity branch of the residual module after FAM. The FSM act before add up to the identity branch. The FSM is placed at the end of each ResNet\cite{he2016deep} block. There are different combination strategy of the LSS and CS block in FSM. In particular, it can combine in serial or parallel. We empirically confirm that exploiting in parallel can greatly improve feature representation ability compared to use the block sequentially. Thus we choose to combine them in parallel.

The FAM applies on the input of the residual bottleneck firstly and the feature selection is acted on the aggregated features. All processes are summarized as follows:
\begin{equation}
\begin{split}
\begin{aligned}
\textbf{F}&=\textbf{A}(\textbf{X}),\\
\textbf{Y} &= \textbf{X}+(\textbf{F}+\alpha\otimes\textbf{F})\oplus(\textbf{F}+\beta\otimes\textbf{F}),\\
\tilde{\textbf{X}} &= \sigma(\textbf{X}+\textbf{Y}),
\end{aligned}
\end{split}
\end{equation}
where the function $\textbf{A}(\textbf{X})$ represents the FAM in the resnet bottleneck, and the $\textbf{F}$ represents the output of FAM from the function $\textbf{A}(\textbf{X})$. The $\oplus$ means the element-wise add between the CS and LSS output. The $\tilde{\textbf{X}}$ is the output feature maps with the enhanced multi-scale discriminative feature information of the human.

\subsection{Feature Fusion Module}
Actually, the spatial-resolution information and semantic concepts can benefit each other. The deeper layers tend to encode high-level semantic information, by contrast, the shallower layers are more likely to capture more spatial features and can better reconstruct spatial information just like the design of the U-Net\cite{ronneberger2015u-net:}. Feature fusion strategy is often employed in multi-person pose estimation in different forms and for different purposes. Many previous methods\cite{lin2017feature,chen2018cascaded,newell2016stacked} combine the low-level and high-level features together. However, since the gap in semantic levels and spatial resolutions, making it's less effective to fuse different level features directly. Our approach is related to the study ExFuse\cite{zhang2018exfuse}. For semantic segmentation, it narrows the gap between spatial resolution and semantic-level features in a different ways thereby progressively improve the segmentation performance. Motivated by this, we propose to take more spatial high-resolution structural feature into semantics and more semantic information into high-resolution spatial features in semantic embedding way to enhance the global context information. In particular, for the spatial structural features, we assign intermediate supervisions to the stages of encoder network to make the spatial features encode more semantic concepts. For the semantic-level features, we involve more spatial information to guide the feature fusion in a semantic embedding way. When we obtain the fused global context information, we can better infer the invisible joints or detangle the interacted joints.

Our FFM is shown as Fig.\ref{fig:2network}, the green dashed line represents taking the semantics into low-level spatial features and the pink dashed line represents taking spatial low-level information into high-level features respectively.

\subsection{Dense Upsampling Convolution}
Multiple approaches can decode more accurate information which is generally lost in the downsampling process. For the pose estimation, different methods use the different upsampling forms to generate the final prediction heatmaps. However, they may not consider that most of the frequently used upsampling operations such as bilinear interpolation, deconvolution, transposed convolution, depth-to-space, bicubic upsampling $\emph{etc}$ cannot recover the detailed information leading to the final performance decrease.

Given an input image $\textbf{X}\in\mathbb{R}^{C \times H \times W}$, the keypoint localization is to generate a heatmap with size $H$$\times$$W$ where each pixel is the coordinate of the location. We feed the image into ResNet\cite{he2016deep} network and output the feature map $\mathbb{R}^{h \times w \times c}$ before making predictions, where $h$ = $\frac{H}{f}$,$w$ = $\frac{W}{f}$, and $f$ is the downsampling factor.
The DUC directly adopts the convolution on the feature maps output by the encoder network and then to generate the pixel-wise prediction map rather than pad zeros in the unpooling step just as the common upsampling methods. Fig.\ref{fig:2network} depicts our model architecture with a DUC module. The DUC is related to convolution operation, which is performed on the feature map output by ResNet\cite{he2016deep} of $\mathbb{R}^{h \times w \times c}$. Then get further output feature map of dimension $\mathbb{R}^{h \times w \times f^{2} \times N }$, where $N$ is the total number of joints in the keypoint localization task. Each layer of the convolution operation learns the prediction for each joint. The output feature map is then reshaped to $\mathbb{R}^{H \times W \times N}$ and apply a softmax function. In the end, we apply an element-wise argmax operation to get the final heatmap. The core of DUC is to divide the whole heatmap into same $f$$^{2}$ sub-parts which have the same height and width as the incoming feature map. That is, we transform the whole heatmap into a smaller map with multiple channels. This operation lets us to apply the convolution operation directly between the input feature map and the output heatmaps without inserting extra values in deconvolutional operation. Consequently, it can avoid some detailed information missing in the process of the upsampling. We can improve the final prediction result by handling with the false alarms intrigued by inaccurate estimations to some samples with those information.

\section{Experiments}
Our multi-person pose estimation framework follows the top-down pipeline. First, we apply a human detector to produce all human bounding boxes in the image. Then for each human bounding box, we apply our proposed network to predict the corresponding human pose. The proposed method is evaluated on two recent standard multi-person datasets and a big new crowded datasets with large interaction and crowd cases: MPII Human Pose benchmark\cite{andriluka20142d}, COCO 2017 Keypoints Challenge dataset\cite{lin2014microsoft} and CrowdPose dataset\cite{li2018crowdpose:}.

\subsection{COCO Keypoint Detection}
\textbf{Datasets and Evaluation metric.} To evaluate the proposed method, we validate our model on the challenging COCO keypoint benchmark\cite{lin2014microsoft}. Our models are only trained on the COCO trainval dataset (includes $57$$K$ images and $150$$K$ person instances) without using any other extra data. Ablation studies are validated on the COCO minival dataset (includes $5$$K$ images). The final results are reported on the COCO test-dev dataset (includes $20$$K$ images) compared with the public state-of-the-art results. We use the standard evaluation metric\cite{lin2014microsoft} that reports the OKS-based AP (average precision) in the experiments, where the OKS (object keypoints similarity) defines the similarity between the predicted heatmap and the groundtruth heatmap.

\textbf {Data Augmentation.} We apply random flip, rotation, and scale in our training stage. The flip value is $0.5$. The scale range is ([$0.7$ $\sim$ $1.3$]), and the random rotation range is ([$-40\degree \sim +40\degree$]).

\textbf {Training.} Our model is implemented in Pytorch\cite{Paszke2017AutomaticDI}. For the training, $2$ NVIDIA GPUs on a server are used. Adam\cite{kingma2015adam:} optimizer is adpoted. The base learning rate is set to $5e$$-$$4$, and is decreased by a factor of $0.1$ at $90$ and $120$ epochs, and finally we train for $150$ epochs. Actually, we train ResNet-$50$ based models take about $96$ hours. The input size of the image is resized to a fixed aspect ratio of height : width = $4 : 3$, e.g., $256$ $\times$ $192$ is used as the default resolution, the same as the SBN\cite{xiao2018simple}.

In this paper, our ResNet backbone is initialized with the weights of the public-released Imagenet\cite{russakovsky2015imagenet} pre-trained model. ResNet\cite{he2016deep} backbones with $50$, $101$ and $152$ layers are experimented. For the different resolutions of the input image (e.g., $256$ $\times$ $192$ and $388$ $\times$ $284$) we also conduct experiments. The backbone ResNet-$50$ and the input size $256$ $\times$ $192$ is used by default.

\textbf {Human Detector and Testing.} We use the same human detector as that in the SBN\cite{xiao2018simple}. It is based on Faster-RCNN\cite{ren2015faster} with mAP of human category $56.4$ AP. During testing, Soft-NMS\cite{Bodla2017Soft} is used to suppress duplicated bounding boxes. As a common practice like\cite{xiao2018simple,chen2018cascaded}, for the flipped image we average the predicted heatmap to get the keypoints location. Each keypoint location is predicted by adjusting the highest heatvalue location with offset from the maximum response to the second largest response.

\begin{figure*}[ht]
\centering
\includegraphics[width=6.8in,height = 1.5in]{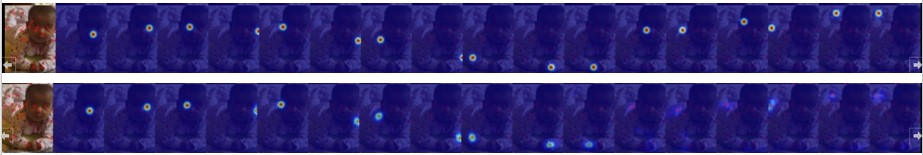}
\caption{Visual heatmaps of SBN\cite{xiao2018simple} and our model on the COCO
minival dataset. From left to right are input images, predicted heatmaps. The first row is generated by our method, the second row is generated by SBN\cite{xiao2018simple}. Best viewed in color}
\label{fig:6visual heatmap}
\end{figure*}

\textbf{Results on COCO val2017.} We compare our model with the stacked Hourglass\cite{newell2016stacked}, CPN\cite{chen2018cascaded} and SBN\cite{xiao2018simple} on the COCO minival dataset as shown in Tab.\ref{tab:11 COCO minival}. The human detection AP of the stacked Hourglass and CPN (57.2\%) is higher than ours (56.4\%). Compared with the $8$-stage Hourglass, our model improves $5.9$ AP, all approaches use an input size of $256$$\times$$192$. Our model outperforms the CPN\cite{chen2018cascaded} and SBN\cite{xiao2018simple} by $3.4$ AP and $2.4$ AP respectively for the input size of $256$ $\times$ $192$ and has an improvement of $2.2$ AP and $1.6$ AP for the input size of $384$ $\times$ $288$.
The qualitative visualization of the improvements between our model and the baseline model is shown in Fig.\ref{fig:6visual heatmap} on the COCO minival dataset.

\begin{table}[htbp]
\centering
\caption{Comparison with the 8-stage Hourglass\cite{newell2016stacked}, CPN\cite{chen2018cascaded}
and SBN\cite{xiao2018simple} on the COCO minival dataset. ¡°*¡± means the model training with
the Online Hard Keypoints Mining.}
\label{tab:11 COCO minival}
\begin{tabular}{c|c|c}
  \toprule
    Models & Input Size & AP(OKS)  \\
     \midrule
    8-stage Hourglass\cite{newell2016stacked} & 256$\times$256 & 67.1 \\
    8-stage Hourglass\cite{newell2016stacked} & 256$\times$192 & 66.9 \\
    CPN\cite{chen2018cascaded} & 256$\times$192 & 68.6 \\
    CPN\cite{chen2018cascaded} & 384$\times$288 & 70.6 \\
    CPN*\cite{chen2018cascaded} & 256$\times$192 & 69.4 \\
    CPN*\cite{chen2018cascaded} & 384$\times$288 & 71.6 \\
    SBN\cite{xiao2018simple} & 256$\times$192 & 70.4 \\
    SBN\cite{xiao2018simple} & 384$\times$288 & 72.2 \\
     \hline

    Ours*(ResNet-50)   & 256$\times$192 & $\textbf{72.8}$  \\
    Ours*(ResNet-50)   & 256$\times$256 & $\textbf{73.1}$  \\
    Ours*(ResNet-50)   & 384$\times$288 & $\textbf{73.8}$  \\
     \bottomrule
  \end{tabular}
   \end{table}

\begin{figure}[h]
\centering
\includegraphics[width=3.0in,height = 2.0in]{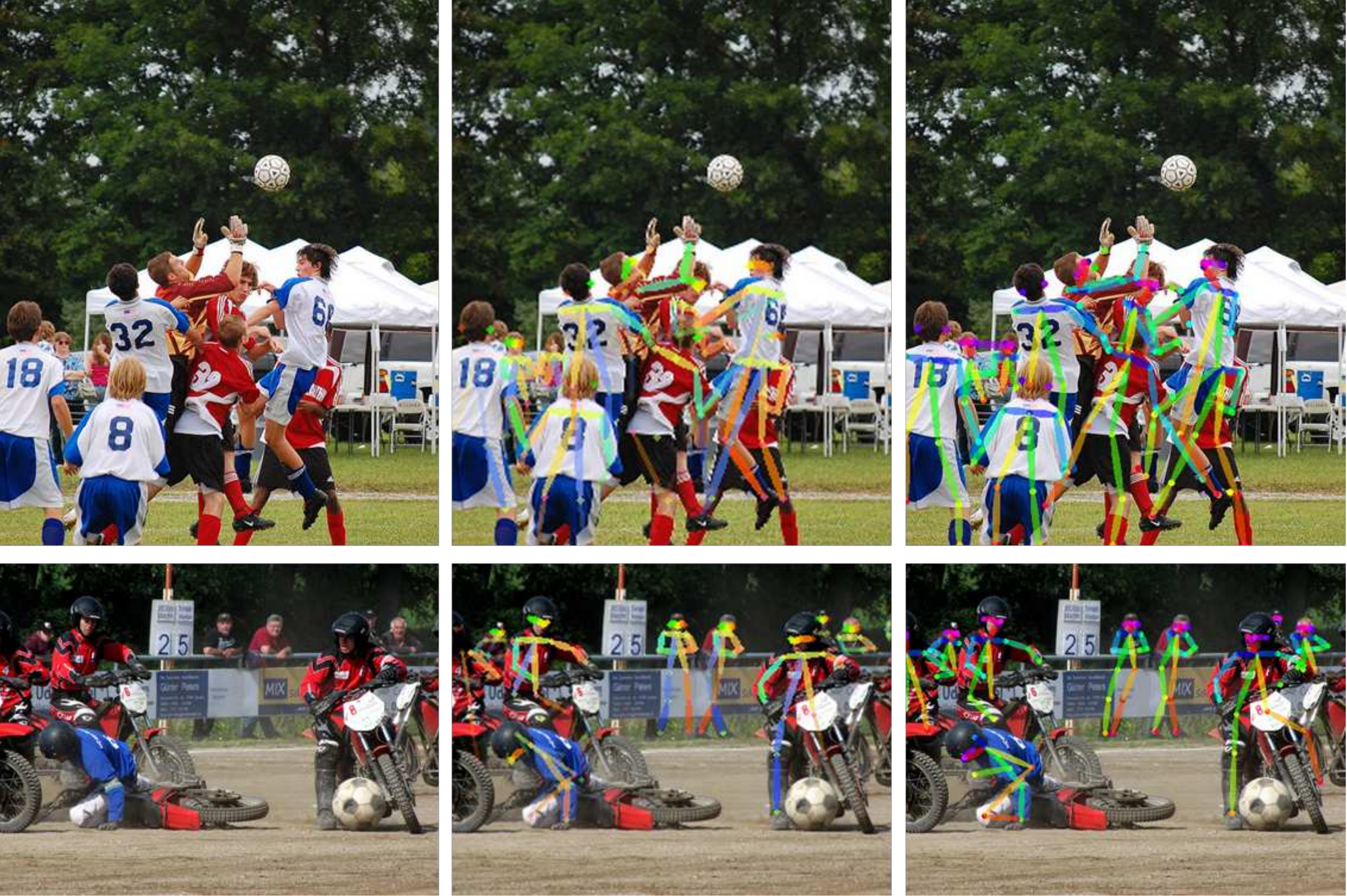}
\caption{Comparison of SBN\cite{xiao2018simple} and our model on the COCO
minival dataset. From left to right are input images, SBN results, ours results.
Best viewed in color.}
\label{fig:7duibi}
\end{figure}

As shown in Fig.\ref{fig:7duibi}, our approach can make better performance for pose estimation, although in challenging cases, ($\emph{e.g.}$, close-interactions in the first row, invisible joints generated by occlusions in the last row) where SBN\cite{xiao2018simple} cannot well deal with.

\begin{figure*}[ht]
\centering
\includegraphics[width=6.8in,height = 2.3in]{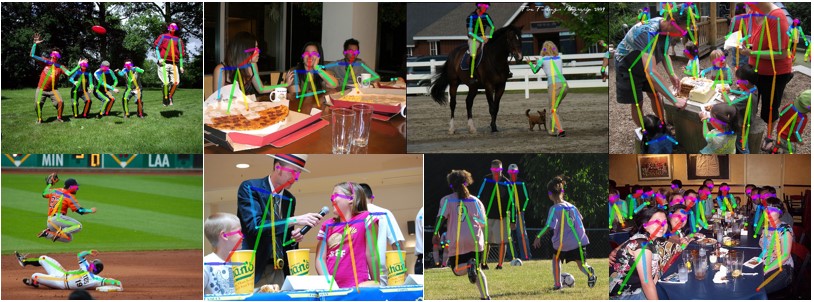}
\caption{Qualitative results of our model on the COCO test-dev dataset. Our model can achieve more accurate result in pose estimation although there exist many problems such as extreme cases (the bottom-left result), complex and crowded poses (the top-right result).}
\label{fig:8coco}
\end{figure*}

\textbf{Results on COCO test-dev 2017.}  Tab.\ref{tab:012 COCO test-dev} demonstrates the results of our method in the test-dev dataset of the COCO dataset. Our approach is obviously better than the existing promising approaches. Additionally, our small network, EFAFNet, achieves an AP of $72.8$. It outperforms almost all the other top-down approaches under the same backbone network, provides comparable results to the latest work HRNet\cite{sun2019deep}. It designs a complex and robust deep high resolution network that can extract the multi-scale features of human better across multi-stage fusions and thus achieves the best performance in the top-down methods. Compared to the SBN\cite{xiao2018simple} with the same input size, our small and big models receive $2.4$ and $1.2$ improvements, respectively. Fig.\ref{fig:8coco} illustrates some qualitative results generated using our method.

\begin{table*}
\centering
\caption{Comparison of final results on the COCO test-dev dataset.$\textbf{Top}$: methods in the literature, trained only with the COCO trainval dataset. $\textbf{Middle}$: results submitted to the COCO test-dev leaderboard\cite{COCO}. ¡°*¡± means that the method involves extra data for the training.¡°+¡± indicates the results using the ensembled models. $\textbf{Bottom}$: the results of our single model, trained only with the COCO trainval dataset.}
\label{tab:012 COCO test-dev}
\begin{tabular}{c|c|c|c|c|c|c|c|c}
\toprule
    Method & Backbone & Input Size & AP & AP .5 & AP .75 & AP (M)& AP (L) & AR  \\
    \midrule
  CMU-Pose\cite{cao2017realtime} &    -     &     -     & 61.8& 84.9 & 67.5  & 57.1 & 68.2  & 66.5 \\
  Mask-RCNN\cite{he2017mask} &ResNet-50-FPN& -     & 63.1& 87.3 & 68.7  & 57.8 & 71.4  &  - \\
Associative Embedding\cite{newell2017associative} &-&512$\times$512& 65.5& 86.8 & 72.3  & 60.6 & 72.6  &  70.2 \\
  G-RMI\cite{papandreou2017towards} &ResNet-101&353$\times$257&64.9& 85.5 & 71.3  & 62.3 & 70.0  &  69.7 \\
  CPN\cite{chen2018cascaded} &ResNet-Inception&384$\times$288&72.1& 91.4 & 80.0  & 68.7 & 77.2  &  78.5 \\
  RMPE\cite{fang2017rmpe:}&PyraNet&320$\times$256&72.3& 89.2 & 79.1  & 68.0 & 78.6  &  - \\
  Integral Pose Regression\cite{sun2018integral}&ResNet-101&256$\times$256&67.8& 88.2 & 74.8& 63.9 & 74.0&- \\
  MultiPoseNet\cite{kocabas2018multiposenet:} &  -   &    -      &69.6& 86.3 & 76.6  & 65.0 & 76.3  &  73.5 \\
  CSANet\cite{2019Pose}&ResNet-101&   384$\times$288  &73.8& 91.7 & 81.4  & 70.4 & 79.6  & 80.3 \\
  SBN\cite{xiao2018simple} &ResNet-101&   384$\times$288  &73.2& 91.4 & 80.9  & 69.7 & 79.5  &  78.5 \\
  SBN\cite{xiao2018simple} &ResNet-152&384$\times$288&73.8& 91.7 & 81.2  & 70.3 & 80.0  &  79.1 \\
  HRNet\cite{sun2019deep} & HRNet-W48 &384$\times$288&$\textbf{75.5}$&
  $\textbf{92.5}$ & $\textbf{83.3}$  & $\textbf{71.9}$ & $\textbf{81.5}$ &  $\textbf{80.5}$ \\
     \hline
   FAIR Mask R-CNN*\cite{COCO} & ResNet-101-FPN    & - & 90.4 & 77.0  & 64.9 & 76.3  & 75.2 \\
    G-RMI*\cite{COCO} & ResNet-152 &353$\times$257& 71.0& 87.9 & 77.7  & 69.0 & 75.2  & 75.8 \\
    oks*\cite{COCO}  &     - &              -  & 72.0& 90.3 & 79.7  & 67.6 & 78.4  & 77.1 \\
    bangbangren*$^{+}$\cite{COCO}& ResNet-101 & -         & 72.8& 89.4 & 79.6  & 68.6 & 80.0  & 78.7 \\
    CPN$^{+}$\cite{COCO}  & ResNet-Inception &384$\times$288& 73.0& 91.7 & 80.9  & 69.5 & 78.1  & 79.0 \\
    HRNet*\cite{sun2019deep} & HRNet-W48 &384$\times$288&$\textbf{77.0}$& $\textbf{92.7}$ & $\textbf{84.5}$  & $\textbf{73.4}$ & $\textbf{83.1}$  &  $\textbf{82.0}$ \\
    \hline
    Ours & ResNet-50 & 256$\times$192  & 72.8& 91.5& 80.5& 69.0& 77.1& 76.4 \\
    Ours & ResNet-50 & 384$\times$288  & 73.8& 91.6& 81.5& 70.7& 79.6& 77.4 \\
    Ours & ResNet-101 & 256$\times$192  & 73.5& 91.6& 81.6& 70.6& 78.2& 77.2 \\
    Ours & ResNet-101 & 384$\times$288  & 74.5& 91.7& 81.9& 71.1& 78.7& 78.6 \\
    Ours & ResNet-152 & 256$\times$192  & 74.2& 91.8& 82.1& 71.3& 79.1& 77.9 \\
    Ours & ResNet-152 & 384$\times$288  & 75.0&92.0& 82.2& 71.6& 80.2& 79.4 \\
   \bottomrule
  \end{tabular}
   \end{table*}

\subsection{MPII Human Pose Estimation}
\textbf{Datasets and Evaluation metric.} The MPII Human Pose dataset\cite{andriluka20142d} consists of images taken from real-world activities with full-body pose annotations. There are about $25$$K$ images with $40$$K$ objects, where there are $12$$K$ objects for testing and the remaining objects for the training set. The data augmentation and the training strategy are the same to COCO, except that the input size is cropped to $256$ $\times$ $256$ for fair comparison with other methods. For evaluation metric, the standard metric PCKh\cite{andriluka20142d} (head-normalized probability of correct keypoint) score is used. The PCKh@$0.5$ score is reported in our result, $50$\% of the head size for normalization.

\textbf{Testing.} The testing procedure is almost the same to that in COCO except that we adopt the standard testing strategy using the provided person boxes instead of detected person boxes.

\textbf{Results on MPII Human Pose.}
Tab.\ref{tab:01 MPII} shows the PCKh@$0.5$ results, we reimplement the SBN\cite{xiao2018simple} by using ResNet-$50$ as the backbone with the input size $256$ $\times$ $256$. Our EFAFNet achieves a $98.2$ PKCh@$0.5$ score, and outperforms the stacked hourglass approach\cite{newell2016stacked} and other methods\cite{chu2017multi-context,ke2018multi-scale,sun2017human,tang2018deeply,yang2017learning}.  We also test our big network ResNet-152 and obtained the best result $98.6$. It proves that out model can get superior performance in this dataset.
\begin{table}[ht]
\caption{Results on MPII test set}
\label{tab:01 MPII}
\centering
\setlength{\tabcolsep}{0.55mm}
\begin{tabular}{c|ccccccc|c}
\toprule
    Method & Head & Shoulder & Elbow & Wrist & Hip & Knee & Ankle & Total \\
    \midrule
    Insafutdinov $\emph{et al.}$\cite{insafutdinov2016deepercut:} & 96.8 & 95.2 & 89.3 & 84.4 & 88.4 & 83.4 & 78.0 & 88.5 \\
    Wei $\emph{et al.}$\cite{wei2016convolutional} & 97.8 & 95.0 & 88.7 & 84.0 & 88.4 & 82.8 & 79.4 & 88.5 \\
    Bulat $\emph{et al.}$\cite{bulat2016human} & 97.9 & 95.1 & 89.9 & 85.3 & 89.4 & 85.7 & 81.7 & 89.7 \\
    Newell $\emph{et al.}$\cite{newell2016stacked} & 98.2 & 96.3& 91.2 & 87.2 & 89.8 & 87.4 & 83.6 & 90.9 \\
    Sun $\emph{et al.}$\cite{sun2017human} & 98.1 & 96.2 & 91.2& 87.2 & 89.8 & 87.4 & 84.1 & 91.0 \\
    Tang $\emph{et al.}$\cite{tang2018quantized} & 97.4 & 96.4 & 92.1 & 87.7 & 90.2 & 87.7 & 84.3 & 91.2 \\
    Ning $\emph{et al.}$\cite{ning2018knowledge-guided} & 98.1 & 96.3 & 92.2 & 87.8 & 90.6 & 87.6 & 82.7 & 91.2 \\
    Luvizon $\emph{et al.}$\cite{luvizon2017human} & 98.1 & 96.6 & 92.0 & 87.5 & 90.6 & 88.0 & 82.7 & 91.2 \\
    Chu   $\emph{et al.}$\cite{chu2017multi-context} & 98.5& 96.3 & 91.9 & 88.1 & 90.6 & 88.0 & 85.0 & 91.5 \\
    Chou $\emph{et al.}$\cite{chou2018self} & 98.2& 96.8 & 92.2 & 88.0 & 91.3 & 89.1 & 84.9 & 91.8 \\
    Chen $\emph{et al.}$\cite{chen2017adversarial} & 98.1 & 96.5 & 92.5 & 88.5 & 90.2 & 89.6 & 86.0 & 91.9 \\
    Yang $\emph{et al.}$\cite{yang2017learning} & 98.5 & 96.7 & 92.5 & 88.7 & 91.1 & 88.6 & 86.0 & 92.0 \\
    Ke $\emph{et al.}$\cite{ke2018multi-scale} & 98.5 & 96.8 & 92.7 & 88.4 & 90.6 & 89.3 & 86.3 & 92.1 \\
    Tang $\emph{et al.}$\cite{tang2018deeply} & 98.4 & 96.9 & 92.6 & 88.7 & 91.8 & 89.4 & 86.2 & 92.3 \\
   \hline
    SBN $\emph{et al.}$\cite{xiao2018simple} & 98.5 & 96.6 & 91.9 & 87.6 & 91.1 & 88.1 & 84.1 & 91.5 \\
    HRNet $\emph{et al.}$\cite{sun2019deep} & \textbf{98.6} & \textbf{96.9} & \textbf{92.8} & \textbf{89.0} & \textbf{91.5} & \textbf{89.0} & \textbf{85.7} & \textbf{92.3} \\
     \hline
    Ours & 98.6 & 96.8 & 92.3 & 88.2 & 91.3 & 88.6 & 85.0 & 92.0 \\
    \bottomrule
  \end{tabular}
   \end{table}

\subsection{CrowdPose}
\textbf{Datasets and Evaluation metric.} CrowdPose dataset\cite{li2018crowdpose:} contains $20$$K$ images in total and $80$$K$ human instances. The CrowdPose dataset divides into three crowding levels by $\emph{Crowd Index}$: easy ($0$ $\sim$ $0.1$), medium ($0.1$ $\sim$ $0.8$) and hard ($0.8$ $\sim$ $1$). Its Crowd Index satisfies uniform distribution in $[0,1]$. CrowdPose dataset aims to promote performance in crowded cases and make models generalize to different scenarios. It uses the same evaluation metric as COCO.

\textbf{Training and Testing.}
The data augmentation and the training strategy are the same as COCO, except that the input size is cropped to $256$ $\times$ $192$ for fair comparison with other methods. The model is only trained on the CrowdPose training set without using any extra data. The testing procedure is almost the same as that in COCO except that the provided human detector of SBN\cite{xiao2018simple}is used for all methods to ensure a fair comparison. We extend $30$\% along the height and width directions of the detected human proposals for guaranteeing that the person body part can be extracted completely .

\textbf{Results on CrowdPose.}
The Tab.\ref{tab:02CrowdPose} show the quantitative results on CrowdPose test set. We can see that our method improves $4$ $\sim$ $5$ mAP compared with most of the advanced methods. It demonstrates the effectiveness of our proposed method in crowded scenes.
\begin{table*}[htbp]
\centering
\caption{Results on CrowdPose test set}
\label{tab:02CrowdPose}
\begin{tabular}{ccccccc}
  \toprule
    Models & mAP @0.5:0.95 & mAP @0.5  & mAP @0.75& mAR @0.5:0.95& mAR @0.5& mAR @0.75\\
  \midrule
    Mask R-CNN\cite{he2017mask} & 57.2  & 83.5& 60.3& 65.9& 89.5& 69.4 \\
    AlphaPose\cite{fang2017rmpe:}& 61.0 & 81.3  & 66.0& 67.6& 86.7& 71.8 \\
    SBN\cite{xiao2018simple}   & 60.8 & 81.4  & 65.7& 67.3& 86.3& 71.8  \\
    CrowdedPose\cite{li2018crowdpose:}   & $\textbf{66.0}$ & $\textbf{84.2}$  & $\textbf{71.5}$& $\textbf{72.7}$& $\textbf{89.5}$& $\textbf{77.5}$  \\
    \hline
    $\textbf{Ours}$ & $\textcolor[rgb]{1.00,0.00,0.00}{64.2}$ & $\textcolor[rgb]{1.00,0.00,0.00}{82.2}$ & $\textcolor[rgb]{1.00,0.00,0.00}{69.5}$& $\textcolor[rgb]{1.00,0.00,0.00}{70.7} $& $\textcolor[rgb]{1.00,0.00,0.00}{87.7}$ & $\textcolor[rgb]{1.00,0.00,0.00}{75.5}$ \\

   \bottomrule
  \end{tabular}
   \end{table*}
We also report the results on three crowding levels as stated before in Tab.\ref{tab:03CrowdPose}, $\emph{i.e.}$, $\textbf{uncrowded}$, $\textbf{medium crowded}$ and $\textbf{extremely crowded}$. It demonstrates that our method get more accurate results across all crowding levels, meanwhile as the scenes are more crowded the relative improvement increases .

\begin{figure*}[ht]
\centering
\includegraphics[width=6.5in,height = 2.6in]{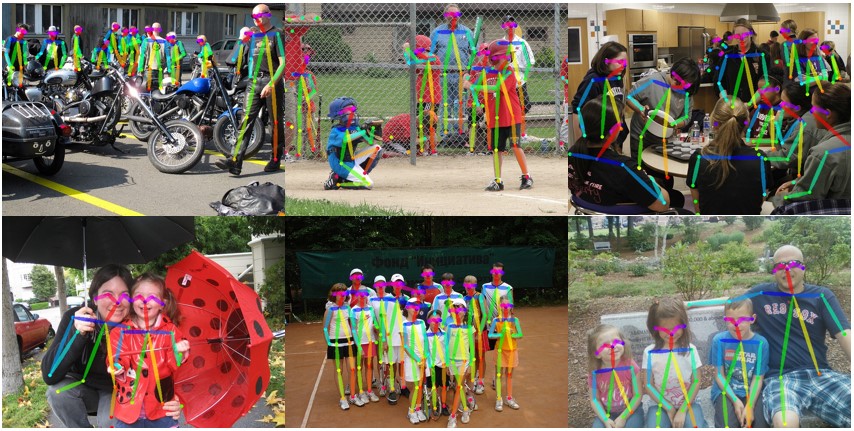}
\caption{Qualitative results of our model on the CrowdPose dataset. Our method can deal with challenges include appearance variation (the top-row), occlusion, clustered (the top-left results), and non-standard poses (the top-right results).}
\label{fig:5crowded}
\end{figure*}

\begin{table}[htbp]
\centering
\caption{Results on CrowdPose test set. Test set is divided into
three parts. FPS column reports the speed on the whole test set. Since the
dataset is quite young, only a few works report results on it}
\label{tab:03CrowdPose}
\setlength{\tabcolsep}{1mm}
\begin{tabular}{c|ccc|c}
  \toprule
    Method & AP$_{easy}$ & AP$_{medium}$ & AP$_{hard}$& FPS \\
      \midrule
    OpenPose\cite{cao2017realtime}& 62.7  & 48.7& 32.3& 5.3 \\
    Mask R-CNN\cite{he2017mask} & 69.4  & 57.9& 45.8& 2.9 \\
    AlphaPose\cite{fang2017rmpe:}& 71.2 & 61.4  & 51.1& 10.9 \\
    SBN\cite{xiao2018simple}   & 71.4 & 61.2  & 51.2& -  \\
    CrowdedPose\cite{li2018crowdpose:}  & $\textbf{75.5}$ & $\textbf{66.3}$  & $\textbf{57.4}$& 10.1  \\
    \hline
    $\textbf{Ours}$& \textcolor[rgb]{1.00,0.00,0.00}{73.6} & \textcolor[rgb]{1.00,0.00,0.00}{63.7}  & \textcolor[rgb]{1.00,0.00,0.00}{55.6}& 7.9 \\
     \bottomrule
  \end{tabular}
   \end{table}
Compared with the baseline SBN\cite{xiao2018simple}, our proposed approach has an significantly improvement and we also provide comparable results to CrowdPose\cite{li2018crowdpose:} across three crowding levels. CrowdPose\cite{li2018crowdpose:} surpasses our final model by $1$ $\sim$ $2$ AP. This can be attributed to their special design for handling crowded scenarios. They add an additional global association algorithm after performing the modified single person pose estimation. Compared with them, our model can be trained in an end-to-end manner without using any extra post-processing method. We achieve comparable results with them in a more efficient way, which can be attributed to our elaborate feature representation and effective decoding. Additionally, there are no other methods reporting experiment result on CrowdPose to our knowledge. The qualitative visualization on CrowdPose dataset is depicted in Fig.\ref{fig:5crowded}.
\subsection{Ablation Study}

In this section, we will decompose our proposed EFAFNet to analyse the effect of each sub module step by step. Moreover, we make all comparisons by experiment on COCO val$2017$ dataset. We adopt the input size $256$ $\times$ $192$ of all our models, and the ResNet-$50$ is adopted for our default backbone.

\subsubsection{FASM: Feature Aggregation and Selection Module}
In this subsection, we will discuss the effect of the FASM. The whole module is made up of two parts: FAM and FSM. We will analyse their effectiveness respectively.

\textbf{Feature Aggragation Module}
Pose estimation is a kind of sparse object detection. The multi-scale representation ability is vital for object detection. Therefore, it's important for the the regression task to possess rich multi-scale representation. We use the SBN\cite{xiao2018simple} as the baseline method, and replace the bottleneck of the backbone network ResNet\cite{he2016deep} with our proposed FAM, while keeping other configurations unchanged. The performance of AP on COCO minival dataset is shown in Tab.\ref{tab:05 ablation}. From the result, we can see that the model with the FAM achieves the AP of $70.9$ AP. It indicates that when only using the FAM, our model outperforms the baseline SBN\cite{xiao2018simple} by $0.6$ AP.

\begin{table}[htbp]
\centering
\caption{Ablation study of each component on the COCO minival dataset.}
\label{tab:05 ablation}
\setlength{\tabcolsep}{1mm}
\begin{tabular}{cccccc}
\toprule
    Method & FAM & FSM & FFM & DUC & AP(OKS)  \\
    \midrule
    SBN\cite{xiao2018simple}(baseline) &  &  & & & 70.4 \\
    SBN\cite{xiao2018simple}+FAM &  $\surd$  &  & &   & 70.9 \\
    SBN\cite{xiao2018simple}+FSM  &  & $\surd$ & & & 71.5 \\
    SBN\cite{xiao2018simple}+FFM   &  & & $\surd$&   & 70.9 \\
    SBN\cite{xiao2018simple}+DUC   &  & & &  $\surd$ & 70.6 \\
    Ours(ensembled)   & $\surd$ &$\surd$ &$\surd$ &$\surd$   & 72.8\\
   \bottomrule
  \end{tabular}
   \end{table}

\textbf{Feature Selection Module}
In this experiment, we explore the effects of different implementation orders of the LSS and the CS block in the residual bottleneck. As shown in Tab.\ref{tab:05 ablation}, the FSM achieves the AP of $71.5$. It indicates that when only using the FSM, our model outperforms the baseline SBN\cite{xiao2018simple} by $1.1$ AP. This demonstrates this module make a big difference for the task. And from Tab.\ref{tab:01 Ablation FSM} we confirm that the best-combining strategy is in parallel, which can further improve the accuracy.

\begin{table}[htbp]
\centering
\caption{Ablation study of the different implementation of CS and LSS on
the COCO minival dataset. CS denotes the Channel-wise Selection
, LSS denotes the Location-sensitive Selection.}
\label{tab:01 Ablation FSM}
\begin{tabular}{cccc}
\toprule
    Method  & AP(OKS)  \\
    \midrule
    SBN\cite{xiao2018simple}(baseline)& 70.4 \\
    SBN\cite{xiao2018simple}+CS+LSS & 70.8 \\
    SBN\cite{xiao2018simple}+LSS+CS & 71.1 \\
    SBN\cite{xiao2018simple}+CS\&LSS in parallel& $\textbf{71.5}$ \\

   \bottomrule
  \end{tabular}
   \end{table}

\begin{figure}[htbp]
\centering
\includegraphics[width=2.8in,height = 2.5in]{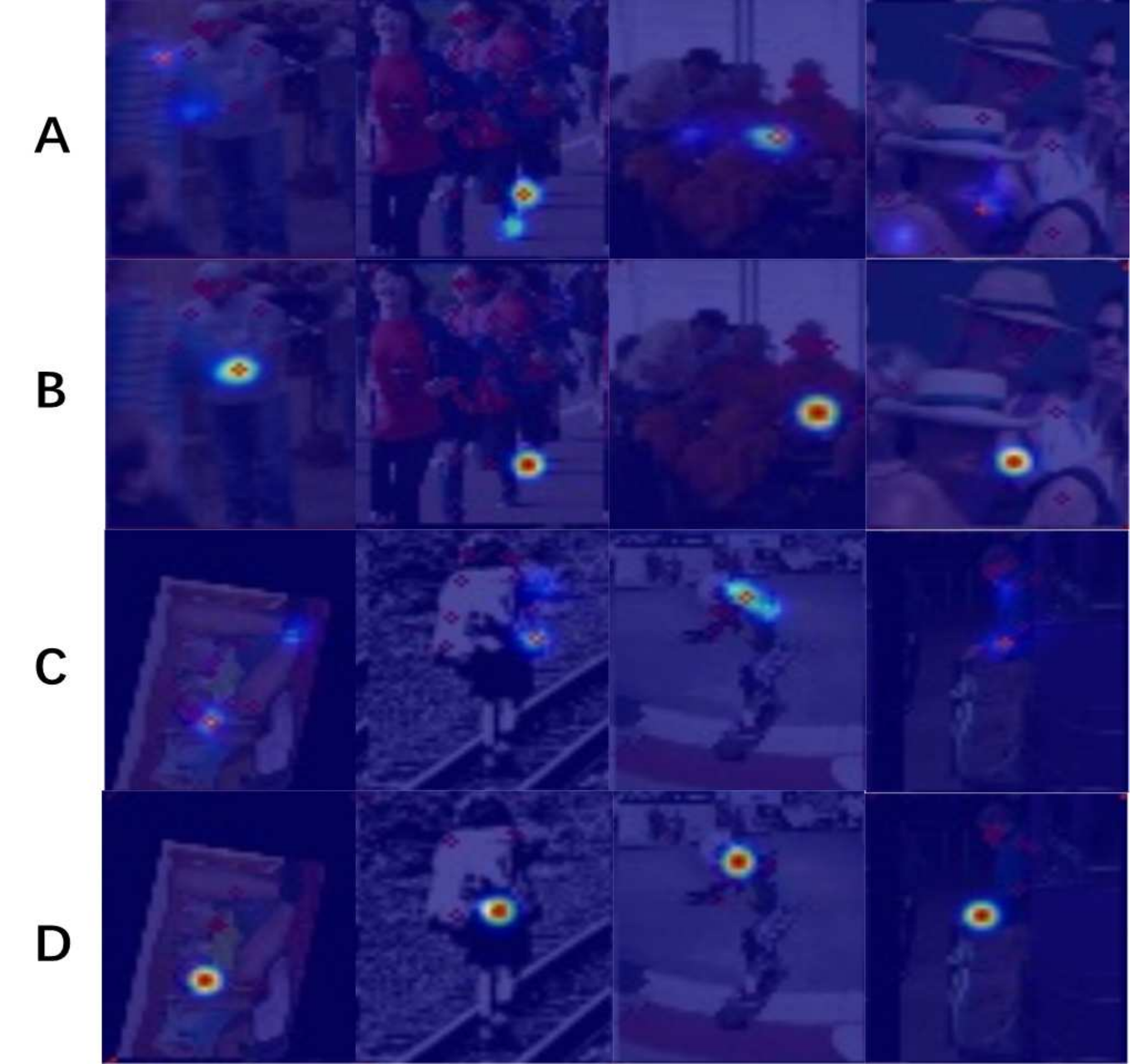}
\caption{Instances of the proposed FSM solve various false alarm challenges occured in previous heatmaps. (A) and (C) show the previous heatmaps of baseline w/o FSM. (B) and (D) show the refined heatmaps of different human body parts through adaptively selecting the discriminative location and channel-wise context information, which can reduce false alarms intrigued by incorrect estimations. The deeper the color corresponds to the larger value of the weight. (Best viewed in color).}
\label{fig:8FSM map}
\end{figure}

To interpret our FSM clearly, we visualize the estimated heatmaps with the FSM as shown in Fig.\ref{fig:8FSM map}. Besides, we observe that FSM can solve many false alarms that happen in previous heatmaps caused by intertwined people, occlusion, and inaccuracy to localize the discriminative body parts more precisely.

\subsubsection{Is Feature Fusion Enhanced ?}
Here we discuss the feature fusion module used in our network. The effect of FFM is shown in Tab.\ref{tab:05 ablation}. The fusion strategy is not always achieve the good performance only if the proper low-level encoder convolution layer concatenates with the high-level decoder layer. The feature maps with different scales extracted from the ResNet\cite{he2016deep} backbone denote as Conv-$2$ $\sim$ $5$. In the decoder stage, we combine the low-level (Conv$2$) feature maps with the high-level (Conv$5$) feature maps together. The proposed EFAFNet makes a considerable result increase of $0.5$ AP, compared with the baseline model. The comparison implies our insights of feature fusion indeed.

\subsubsection{Dense Upsampling Convolution}
Here, we analyse the effect of DUC based on baseline framework. In this module, we can change the dimension of the top ResNet output feature maps. We assume that the dimension of the output layer is $64$ $\times$ $64$ $\times$ $17$ in the baseline model ($17$ is the joint number), and for the same layer with DUC dimension is $64$ $\times$ $64$ $\times$ ($r$$^{2}$ $\times$ $17$) where the $r$ is the downsampling rate of the model ($r$ = $8$ in our paper). We reshape the predicted heatmap to $512$ $\times$ $512$ $\times$ $17$. This module increase few parameters compared to the previous model. When we train this model of DUC, we adopt the similar set with the baseline model and train for $50$ epochs, and finally achieve a AP of $70.6$ on the COCO minival dataset as shown in Tab\ref{tab:05 ablation}, a $0.2$ AP increase compared to the baseline model.
\subsubsection{Bottleneck: The effective location to apply FASM}
We empirically exploit the most effective location to apply our FASM is the bottleneck of the model. The previous work SE-Net\cite{hu2018squeeze-and-excitation} mostly concentrate on making their contribution in the $\emph{¡®convolution blocks¡¯}$ instead of the $\emph{¡®bottlenecks¡¯}$. By comparing different model performance in different locations on COCO minival dataset. We can conclude that applying the FASM at the bottleneck can achieve best result considering the speed and accuracy trade-offs just as shown in Tab.\ref{tab:02 Bottleneck location}. In most cases, we achieve better accuracy at a little overhead cost.

\begin{table}[htbp]
\centering
\caption{Comparison of FASM in different points. FASM-C denotes where the module is inserted to convolution block.}
\label{tab:02 Bottleneck location}
\begin{tabular}{cccc}
  \toprule
    Model & Params & GFLOPs & AP  \\
    \midrule
    SBN\cite{xiao2018simple} & 34.0M & 8.90 & 70.4  \\
    SBN\cite{xiao2018simple}+FASM-C & 39.28M & 10.52 & 70.8  \\
    SBN\cite{xiao2018simple}+FASM(Our) & 37.0M & 9.23 & $\textbf{71.5}$  \\
    \bottomrule
  \end{tabular}
   \end{table}

\subsubsection{Human Detection Performance}
Tab.\ref{tab:04 human detection} shows the relationships between the human detection result and the corresponding pose estimation performance on the COCO minival dataset. Our model and SBN\cite{xiao2018simple} are compared in this experiment. Both models are trained with the ResNet-$50$ backbone and the $256$ $\times$ $192$ input size. The SBN\cite{xiao2018simple} adopts the Faster-RCNN\cite{ren2015faster} as the human detector, which reports the human detection AP $56.4$ in their paper. We adopt the same human detector with them for fair comparison. Additionally, we also use the human detection results provided by the CPN\cite{chen2018cascaded} (reports the human detection AP $57.2$) for comparison.
\begin{table}[htbp]
\centering
\caption{Comparison of pose estimation methods with different human detectors on the COCO minival dataset. All pose estimation methods are trained with the ResNet-50 backbone and the 256$\times$192 input size.}
\label{tab:04 human detection}
\begin{tabular}{cccc}
\toprule
    Pose Method & Det Method & Human AP & Pose AP \\
 \midrule
    SBN\cite{xiao2018simple} & Faster-Rcnn\cite{ren2015faster} & 56.4 & 70.4  \\
    CPN\cite{chen2018cascaded} &  - \cite{chen2018cascaded} & 57.2 & 69.4  \\
    RMPE\cite{fang2017rmpe:} & SSD-512\cite{liu2016ssd:} & 55.5 & 70.3  \\
    \hline
     Ours & Faster-Rcnn\cite{ren2015faster} & 56.4 & $\textbf{72.8}$  \\
    Ours & - \cite{chen2018cascaded} & 57.2 & $\textbf{73.5}$  \\

    \bottomrule
  \end{tabular}
   \end{table}

From the Tab.\ref{tab:04 human detection} we can see that the pose estimation AP gains increasingly when the human detection AP increases. For example, when the human detection AP increases from $56.4$ to $57.2$, the pose estimation AP of our model increases from $72.8$ to $73.5$. However, although the human detection AP $57.2$ of the CPN\cite{chen2018cascaded} (besides the human detection AP $62.9$ of the CPN\cite{chen2018cascaded} depicted on the COCO test-dev dataset) is higher than ours $56.4$ AP, the pose estimation AP $69.4$ of the CPN\cite{chen2018cascaded} is lower than ours $72.8$ AP. It shows our pose estimator is superior even when equipped with less accurate human detector.

\subsubsection{Online Hard Keypoints Mining}
We examine the loss used in our network. In particular, the mean squared error (MSE) is adopted as out loss function of the model to compare the predicted heatmaps and the groundtruth heatmaps difference. We adopt $2$D Gaussian with standard deviation of $1$ pixel centered on the groundtruth location of each keypoint to produce the groundtruth heatmaps. The OHKM strategy is only calculate the top $R$ ($R$ $\textless$ $K$) keypoints losses of $K$ ( $K$ is the total number of labeled keypoints, say $17$ in this case). The effect of value $R$ is shown in Tab.\ref{tab:05 OHKM}. We achieve the best performance when $R$$=$$8$. Therefore, we set the value of $R$ is $8$ in this case.

\begin{table}[htbp]
\centering
\caption{Comparison of models of hard keypoints numbers in online
hard keypoints mining strategy}
\label{tab:05 OHKM}

\begin{tabular}{|c|c|c|c|c|c|c|}
\hline
    R & 7 & 8 & 11& 13 & 15 & 16 \\
    \hline
    $\textbf{AP(OKS)}$ & 71.3 & $\textbf{72.8}$ & 71.2& 71.6 & 71.2 & 72.3 \\
   \hline
  \end{tabular}
   \end{table}

\subsubsection{Effect of Data Pre-processing}
Here, we discuss the effect of the input sizes of our EFAFNet. All methods use the same backbone of the ResNet-$50$. As Tab.\ref{tab:06 input size} illustrates, we can find that our model outperforms the SBN\cite{xiao2018simple} by $2.4$ AP when the input size of $256$ $\times$ $192$. The AP increase $1.6$ when the input size is $384$ $\times$ $288$.

\begin{table}[htbp]
\setlength{\abovecaptionskip}{0pt}

\centering
\caption{Effect of input size of the image.¡°*¡± means the model training with the Online Hard Keypoints Mining.}
\label{tab:06 input size}
\begin{tabular}{c|c|c}
\toprule
    Models & Input Size & AP(OKS)  \\
     \midrule
    SBN\cite{xiao2018simple} & 256$\times$192 & 70.4 \\
    SBN\cite{xiao2018simple} & 384$\times$288 & 72.2 \\
     \hline
     Ours*(ResNet-50)   & 256$\times$192 & $\textbf{72.8}$  \\
    Ours*(ResNet-50)   & 256$\times$256 & $\textbf{73.1}$  \\
    Ours*(ResNet-50)   & 384$\times$288 & $\textbf{73.8}$  \\
   \bottomrule
  \end{tabular}
   \end{table}
\subsubsection{Effect of Backbone Network}
As we all know, the backbone model become deeper, the performance can be better. We use the ResNet-$50$, ResNet-$101$, and ResNet-$152$ backbones conduct experiments with the input image size of $384$ $\times$ $288$. Tab.\ref{tab:07 Backbone} shows that the backbone change from ResNet-$50$ to ResNet-$101$ and ResNet-$152$ the AP increase is $0.7$ and $1.2$ respectively.
\begin{table}[htbp]
\centering
\caption{Comparison of models of different backbone}
\label{tab:07 Backbone}
\begin{tabular}{c|c|c}
\toprule
    Models & Input Size & AP(OKS)  \\
    \midrule
    8-stage Hourglass\cite{newell2016stacked} & 256$\times$192 & 66.9 \\
    Ours(ResNet-50)   & 384$\times$288 & $\textbf{73.8}$ \\
    Ours(ResNet-101)   & 384$\times$288& $\textbf{74.5}$  \\
    Ours(ResNet-152)   & 384$\times$288 & $\textbf{75.0}$  \\
   \bottomrule
  \end{tabular}
   \end{table}

\section{Conclusion}

In this paper, we handle the multi-person pose estimation with the top-down pipeline. The Feature Aggregation and Selection Module (FASM) embedded in residual bottleneck is designed to adaptively highlight the discrimination of the multi-scale features, for precise joint location and rich spatial context. The Feature Fusion (FF) strategy bridges the gap between semantic information and spatial structural features in semantic embedding to make them reinforce each other. The Dense Upsampling Convolution (DUC) module is proposed to produce elaborate prediction on feature maps to help recover detailed keypoint information. Overall, our model achieves outstanding performance consistently on popular multi-person pose estimation datasets, especially a dataset in crowded scenes.

\section*{Acknowledgment}

This research is supported by National Nature Science Foundation of China under Grants (No.61473031), the Fundamental Research Funds for the Central Universities (2019JBM019)

\ifCLASSOPTIONcaptionsoff
  \newpage
\fi

{
\bibliographystyle{plain}
\bibliography{egbib}

\begin{thebibliography}{10}

\bibitem{andriluka20142d}
Mykhaylo Andriluka, Leonid Pishchulin, Peter~V Gehler, and Bernt Schiele.
\newblock 2d human pose estimation: New benchmark and state of the art
  analysis.
\newblock {\em Computer Vision and Pattern Recognition}, pages 3686--3693,
  2014.

\bibitem{Bodla2017Soft}
Navaneeth Bodla, Bharat Singh, Rama Chellappa, and Larry~S. Davis.
\newblock Soft-nms ¡ª improving object detection with one line of code.
\newblock {\em International Conference on Computer Vision}, pages 5562--5570,
  2017.

\bibitem{Buitelaar2018MixedEmotions}
Paul Buitelaar, Ian Wood, Sapna Negi, Mihael Arcan, John~P. Mccrae, Andrejs
  Abele, Cecile Robin, Vladimir Andryushechkin, Hesam Sagha, and Maximilian
  Schmitt.
\newblock Mixedemotions: An open-source toolbox for multi-modal emotion
  analysis.
\newblock {\em IEEE Transactions on Multimedia}, pages 1--1, 2018.

\bibitem{bulat2016human}
Adrian Bulat and Georgios Tzimiropoulos.
\newblock Human pose estimation via convolutional part heatmap regression.
\newblock {\em European Conference on Computer Vision}, pages 717--732, 2016.

\bibitem{cai2016effective}
Xingyang Cai, Wengang Zhou, Lei Wu, Jiebo Luo, and Houqiang Li.
\newblock Effective active skeleton representation for low latency human action
  recognition.
\newblock {\em IEEE Transactions on Multimedia}, pages 141--154, 2016.

\bibitem{cao2017realtime}
Zhe Cao, Tomas Simon, Shihen Wei, and Yaser Sheikh.
\newblock Realtime multi-person 2d pose estimation using part affinity fields.
\newblock {\em Computer Vision and Pattern Recognition}, pages 1302--1310,
  2017.

\bibitem{chen2018deeplab:}
Liangchieh Chen, George Papandreou, Iasonas Kokkinos, Kevin~P Murphy, and
  Alan~L Yuille.
\newblock Deeplab: Semantic image segmentation with deep convolutional nets,
  atrous convolution, and fully connected crfs.
\newblock {\em IEEE Transactions on Pattern Analysis and Machine Intelligence},
  pages 834--848, 2018.

\bibitem{chen2018cascaded}
Yilun Chen, Zhicheng Wang, Yuxiang Peng, Zhiqiang Zhang, Gang Yu, and Jian Sun.
\newblock Cascaded pyramid network for multi-person pose estimation.
\newblock {\em Computer Vision and Pattern Recognition}, pages 7103--7112,
  2018.

\bibitem{chen2017adversarial}
Yu~Chen, Chunhua Shen, Xiushen Wei, Lingqiao Liu, and Jian Yang.
\newblock Adversarial posenet: A structure-aware convolutional network for
  human pose estimation.
\newblock {\em International Conference on Computer Vision}, pages 1221--1230,
  2017.

\bibitem{chou2018self}
Chiajung Chou, Juiting Chien, and Hwanntzong Chen.
\newblock Self adversarial training for human pose estimation.
\newblock {\em Asia Pacific Signal and Information Processing Association
  Annual Summit and Conference}, pages 17--30, 2018.

\bibitem{chu2017multi-context}
Xiao Chu, Wei Yang, Wanli Ouyang, Cheng Ma, Alan~L Yuille, and Xiaogang Wang.
\newblock Multi-context attention for human pose estimation.
\newblock {\em Computer Vision and Pattern Recognition}, pages 5669--5678,
  2017.

\bibitem{fang2017rmpe:}
Haoshu Fang, Shuqin Xie, Yuwing Tai, and Cewu Lu.
\newblock Rmpe: Regional multi-person pose estimation.
\newblock {\em International Conference on Computer Vision}, pages 2353--2362,
  2017.

\bibitem{he2017mask}
Kaiming He, Georgia Gkioxari, Piotr Dollar, and Ross~B Girshick.
\newblock Mask r-cnn.
\newblock {\em International Conference on Computer Vision}, pages 2980--2988,
  2017.

\bibitem{he2016deep}
Kaiming He, Xiangyu Zhang, Shaoqing Ren, and Jian Sun.
\newblock Deep residual learning for image recognition.
\newblock {\em Computer Vision and Pattern Recognition}, pages 770--778, 2016.

\bibitem{hou2017deeply}
Qibin Hou, Mingming Cheng, Xiaowei Hu, Ali Borji, Zhuowen Tu, and Philip H~S
  Torr.
\newblock Deeply supervised salient object detection with short connections.
\newblock {\em Computer Vision and Pattern Recognition}, pages 815--828, 2017.

\bibitem{hu2018squeeze-and-excitation}
Jie Hu, Li~Shen, and Gang Sun.
\newblock Squeeze-and-excitation networks.
\newblock {\em Computer Vision and Pattern Recognition}, pages 7132--7141,
  2018.

\bibitem{insafutdinov2016deepercut:}
Eldar Insafutdinov, Leonid Pishchulin, Bjoern Andres, Mykhaylo Andriluka, and
  Bernt Schiele.
\newblock Deepercut: A deeper, stronger, and faster multi-person pose
  estimation model.
\newblock {\em European Conference on Computer Vision}, pages 34--50, 2016.

\bibitem{jain2014learning}
Arjun Jain, Jonathan Tompson, Mykhaylo Andriluka, Graham~W Taylor, and
  Christoph Bregler.
\newblock Learning human pose estimation features with convolutional networks.
\newblock {\em International Conference on Learning Representations}, pages
  1--14, 2014.

\bibitem{kadu2014automatic}
Harshad Kadu and C~C~Jay Kuo.
\newblock Automatic human mocap data classification.
\newblock {\em IEEE Transactions on Multimedia}, pages 2191--2202, 2014.

\bibitem{ke2018multi-scale}
Lipeng Ke, Mingching Chang, Honggang Qi, and Siwei Lyu.
\newblock Multi-scale structure-aware network for human pose estimation.
\newblock {\em European Conference on Computer Vision}, pages 731--746, 2018.

\bibitem{kingma2015adam:}
Diederik~P Kingma and Jimmy Ba.
\newblock Adam: A method for stochastic optimization.
\newblock {\em International Conference on Learning Representations}, 2015.

\bibitem{kocabas2018multiposenet:}
Muhammed Kocabas, Salih Karagoz, and Emre Akbas.
\newblock Multiposenet: Fast multi-person pose estimation using pose residual
  network.
\newblock {\em European Conference on Computer Vision}, pages 437--453, 2018.

\bibitem{li2018crowdpose:}
Jiefeng Li, Can Wang, Hao Zhu, Yihuan Mao, Haoshu Fang, and Cewu Lu.
\newblock Crowdpose: Efficient crowded scenes pose estimation and a new
  benchmark.
\newblock {\em Computer Vision and Pattern Recognition}, 2019.

\bibitem{li2019multi-person}
Miaopeng Li, Zimeng Zhou, and Xinguo Liu.
\newblock Multi-person pose estimation using bounding box constraint and lstm.
\newblock {\em IEEE Transactions on Multimedia}, pages 1--1, 2019.

\bibitem{lin2017feature}
Tsungyi Lin, Piotr Dollar, Ross~B Girshick, Kaiming He, Bharath Hariharan, and
  Serge~J Belongie.
\newblock Feature pyramid networks for object detection.
\newblock {\em Computer Vision and Pattern Recognition}, pages 936--944, 2017.

\bibitem{lin2014microsoft}
Tsungyi Lin, Michael Maire, Serge~J Belongie, James Hays, Pietro Perona, Deva
  Ramanan, Piotr Dollar, and C~Lawrence Zitnick.
\newblock Microsoft coco: Common objects in context.
\newblock {\em European Conference on Computer Vision}, pages 740--755, 2014.

\bibitem{liu2016ssd:}
Wei Liu, Dragomir Anguelov, Dumitru Erhan, Christian Szegedy, Scott~E Reed,
  Chengyang Fu, and Alexander~C Berg.
\newblock Ssd: Single shot multibox detector.
\newblock {\em European Conference on Computer Vision}, pages 21--37, 2016.

\bibitem{long2015fully}
Jonathan Long, Evan Shelhamer, and Trevor Darrell.
\newblock Fully convolutional networks for semantic segmentation.
\newblock {\em Computer Vision and Pattern Recognition}, pages 3431--3440,
  2015.

\bibitem{lowe2004distinctive}
David~G Lowe.
\newblock Distinctive image features from scale-invariant keypoints.
\newblock {\em International Journal of Computer Vision}, pages 91--110, 2004.

\bibitem{luvizon2017human}
Diogo~C Luvizon, Hedi Tabia, and David Picard.
\newblock Human pose regression by combining indirect part detection and
  contextual information.
\newblock {\em Computer Vision and Pattern Recognition}, 2017.

\bibitem{marcosramiro2015let}
Alvaro Marcosramiro, Daniel Pizarro, Marta Marronromera, and Daniel
  Gaticaperez.
\newblock Let your body speak: Communicative cue extraction on natural
  interaction using rgbd data.
\newblock {\em IEEE Transactions on Multimedia}, pages 1721--1732, 2015.

\bibitem{COCO}
MS-COCO.
\newblock Coco keypoint leaderboard.
\newblock \url{http://cocodataset.org/}.

\bibitem{newell2017associative}
Alejandro Newell, Zhiao Huang, and Jia Deng.
\newblock Associative embedding: End-to-end learning for joint detection and
  grouping.
\newblock {\em Neural Information Processing Systems}, pages 2277--2287, 2017.

\bibitem{newell2016stacked}
Alejandro Newell, Kaiyu Yang, and Jia Deng.
\newblock Stacked hourglass networks for human pose estimation.
\newblock {\em European Conference on Computer Vision}, pages 483--499, 2016.

\bibitem{ning2018knowledge-guided}
Guanghan Ning, Zhi Zhang, and Zhiquan He.
\newblock Knowledge-guided deep fractal neural networks for human pose
  estimation.
\newblock {\em IEEE Transactions on Multimedia}, pages 1246--1259, 2018.

\bibitem{noh2015learning}
Hyeonwoo Noh, Seunghoon Hong, and Bohyung Han.
\newblock Learning deconvolution network for semantic segmentation.
\newblock {\em International Conference on Computer Vision}, pages 1520--1528,
  2015.

\bibitem{papandreou2017towards}
George Papandreou, Tyler Zhu, Nori Kanazawa, Alexander Toshev, Jonathan
  Tompson, Chris Bregler, and Kevin~P Murphy.
\newblock Towards accurate multi-person pose estimation in the wild.
\newblock {\em Computer Vision and Pattern Recognition}, pages 3711--3719,
  2017.

\bibitem{Paszke2017AutomaticDI}
Adam Paszke, Sam Gross, Soumith Chintala, Gregory Chanan, Edward Yang, Zachary
  DeVito, Zeming Lin, Alban Desmaison, Luca Antiga, and Adam Lerer.
\newblock Automatic differentiation in pytorch.
\newblock {\em Neural Information Processing Systems}, 2017.

\bibitem{pishchulin2013poselet}
Leonid Pishchulin, Micha Andriluka, Peter~V Gehler, and Bernt Schiele.
\newblock Poselet conditioned pictorial structures.
\newblock {\em Computer Vision and Pattern Recognition}, pages 588--595, 2013.

\bibitem{pishchulin2016deepcut:}
Leonid Pishchulin, Eldar Insafutdinov, Siyu Tang, Bjoern Andres, Mykhaylo
  Andriluka, Peter~V Gehler, and Bernt Schiele.
\newblock Deepcut: Joint subset partition and labeling for multi person pose
  estimation.
\newblock {\em Computer Vision and Pattern Recognition}, pages 4929--4937,
  2016.

\bibitem{ren2012visual}
Reede Ren and John~P Collomosse.
\newblock Visual sentences for pose retrieval over low-resolution cross-media
  dance collections.
\newblock {\em IEEE Transactions on Multimedia}, pages 1652--1661, 2012.

\bibitem{ren2015faster}
Shaoqing Ren, Kaiming He, Ross~B Girshick, and Jian Sun.
\newblock Faster r-cnn: towards real-time object detection with region proposal
  networks.
\newblock {\em Neural Information Processing Systems}, pages 91--99, 2015.

\bibitem{ronneberger2015u-net:}
Olaf Ronneberger, Philipp Fischer, and Thomas Brox.
\newblock U-net: Convolutional networks for biomedical image segmentation.
\newblock {\em Medical Image Computing and Computer Assisted Intervention},
  pages 234--241, 2015.

\bibitem{russakovsky2015imagenet}
Olga Russakovsky, Jia Deng, Hao Su, Jonathan Krause, Sanjeev Satheesh, Sean Ma,
  Zhiheng Huang, Andrej Karpathy, Aditya Khosla, Michael~S Bernstein, et~al.
\newblock Imagenet large scale visual recognition challenge.
\newblock {\em International Journal of Computer Vision}, pages 211--252, 2015.

\bibitem{sapp2013modec:}
Benjamin Sapp and Ben Taskar.
\newblock Modec: Multimodal decomposable models for human pose estimation.
\newblock {\em Computer Vision and Pattern Recognition}, pages 3674--3681,
  2013.

\bibitem{2019Pose}
Yu~Su and Wang Xu.
\newblock Multi-person pose estimation with enhanced channel-wise and spatial
  information.
\newblock {\em Computer Vision and Pattern Recognition}, 2019.

\bibitem{sun2017human}
Ke~Sun, Cuiling Lan, Junliang Xing, Wenjun Zeng, Dong Liu, and Jingdong Wang.
\newblock Human pose estimation using global and local normalization.
\newblock {\em International Conference on Computer Vision}, pages 5600--5608,
  2017.

\bibitem{sun2019deep}
Ke~Sun, Bin Xiao, Dong Liu, and Jingdong Wang.
\newblock Deep high-resolution representation learning for human pose
  estimation.
\newblock {\em Computer Vision and Pattern Recognition}, pages 5693--5703,
  2019.

\bibitem{sun2011articulated}
Min Sun and Silvio Savarese.
\newblock Articulated part-based model for joint object detection and pose
  estimation.
\newblock {\em International Conference on Computer Vision}, pages 723--730,
  2011.

\bibitem{sun2018integral}
Xiao Sun, Bin Xiao, Fangyin Wei, Shuang Liang, and Yichen Wei.
\newblock Integral human pose regression.
\newblock {\em European Conference on Computer Vision}, pages 536--553, 2018.

\bibitem{szegedy2015going}
Christian Szegedy, Wei Liu, Yangqing Jia, Pierre Sermanet, Scott~E Reed,
  Dragomir Anguelov, Dumitru Erhan, Vincent Vanhoucke, and Andrew Rabinovich.
\newblock Going deeper with convolutions.
\newblock {\em Computer Vision and Pattern Recognition}, pages 1--9, 2015.

\bibitem{tang2018deeply}
Wei Tang, Pei Yu, and Ying Wu.
\newblock Deeply learned compositional models for human pose estimation.
\newblock {\em European Conference on Computer Vision}, pages 197--214, 2018.

\bibitem{tang2018quantized}
Zhiqiang Tang, Xi~Peng, Shijie Geng, Lingfei Wu, Shaoting Zhang, and Dimitris~N
  Metaxas.
\newblock Quantized densely connected u-nets for efficient landmark
  localization.
\newblock {\em European Conference on Computer Vision}, pages 348--364, 2018.

\bibitem{tian2012exploring}
Yuandong Tian, C~Lawrence Zitnick, and Srinivasa~G Narasimhan.
\newblock Exploring the spatial hierarchy of mixture models for human pose
  estimation.
\newblock {\em European Conference on Computer Vision}, pages 256--269, 2012.

\bibitem{tompson2014joint}
Jonathan Tompson, Arjun Jain, Yann Lecun, and Christoph Bregler.
\newblock Joint training of a convolutional network and a graphical model for
  human pose estimation.
\newblock {\em Neural Information Processing Systems}, pages 1799--1807, 2014.

\bibitem{torres2018a}
Carlos Torres, Jeffrey Fried, Kenneth Rose, and B~S Manjunath.
\newblock A multiview multimodal system for monitoring patient sleep.
\newblock {\em IEEE Transactions on Multimedia}, pages 3057--3068, 2018.

\bibitem{toshev2014deeppose:}
Alexander Toshev and Christian Szegedy.
\newblock Deeppose: Human pose estimation via deep neural networks.
\newblock {\em Computer Vision and Pattern Recognition}, pages 1653--1660,
  2014.

\bibitem{wang2018understanding}
Panqu Wang, Pengfei Chen, Ye~Yuan, Ding Liu, Zehua Huang, Xiaodi Hou, and
  Garrison~W Cottrell.
\newblock Understanding convolution for semantic segmentation.
\newblock {\em Workshop on Applications of Computer Vision}, pages 1451--1460,
  2018.

\bibitem{wei2016convolutional}
Shihen Wei, Varun Ramakrishna, Takeo Kanade, and Yaser Sheikh.
\newblock Convolutional pose machines.
\newblock {\em Computer Vision and Pattern Recognition}, pages 4724--4732,
  2016.

\bibitem{xiao2018simple}
Bin Xiao, Haiping Wu, and Yichen Wei.
\newblock Simple baselines for human pose estimation and tracking.
\newblock {\em European Conference on Computer Vision}, pages 472--487, 2018.

\bibitem{yang2017learning}
Wei Yang, Shuang Li, Wanli Ouyang, Hongsheng Li, and Xiaogang Wang.
\newblock Learning feature pyramids for human pose estimation.
\newblock {\em International Conference on Computer Vision}, pages 1290--1299,
  2017.

\bibitem{zeiler2014visualizing}
Matthew~D Zeiler and Rob Fergus.
\newblock Visualizing and understanding convolutional networks.
\newblock {\em European Conference on Computer Vision}, pages 818--833, 2014.

\bibitem{zhang2018exfuse}
Zhenli Zhang, Xiangyu Zhang, Chao Peng, Xiangyang Xue, and Jian Sun.
\newblock Exfuse: Enhancing feature fusion for semantic segmentation.
\newblock {\em European Conference on Computer Vision}, pages 273--288, 2018.

\bibitem{ZhaoxuanAttention}
Fan Zhaoxuan, Zhao Xu, Lin Tianwei, and Su~Haisheng.
\newblock Attention based multi-view re-observation fusion network for skeletal
  action recognition.
\newblock {\em IEEE Transactions on Multimedia}, pages 363--374, 2018.

\bibitem{zhou2016spatio-temporal}
Feng Zhou and Fernando~De La~Torre.
\newblock Spatio-temporal matching for human pose estimation in video.
\newblock {\em IEEE Transactions on Pattern Analysis and Machine Intelligence},
  pages 1492--1504, 2016.

\end{thebibliography}
}
\end{document}